%% file: ms.tex
\documentclass[preprint]{elsarticle}
\usepackage[utf8]{inputenc}
\usepackage{multirow}
\usepackage{amsfonts}
\usepackage{amssymb}
\usepackage{amsmath}
\usepackage{pgfplots}
\usepackage[lofdepth,lotdepth]{subfig}
\usepackage{adjustbox}
\usepackage{booktabs}
\usepackage{url}
\usepackage{ulem}

\newcommand{\marius}[1]{#1}

\usetikzlibrary{positioning, fit, calc, arrows.meta}

\DeclareMathOperator*{\argmax}{arg\,max}
\DeclareMathOperator*{\argmin}{arg\,min}

\biboptions{colon}
\journal{}

\title{Unsupervised Visual Feature Learning with Spike-timing-dependent Plasticity: How Far are we from Traditional Feature Learning Approaches?}

\author[univlille]{Pierre Falez}
\ead{pierre.falez@univ-lille.fr}

\author[univlilleimt]{Pierre Tirilly\corref{cor1}}
\ead{pierre.tirilly@univ-lille.fr}

\author[univlille]{Ioan Marius Bilasco}
\ead{marius.bilasco@univ-lille.fr}

\author[univlille]{Philippe Devienne}
\ead{philippe.devienne@univ-lille.fr}

\author[univlille]{Pierre Boulet}
\ead{pierre.boulet@univ-lille.fr}

\cortext[cor1]{Corresponding author}

\address[univlille]{Univ. Lille, CNRS, Centrale Lille, UMR 9189 - CRIStAL - Centre de Recherche en Informatique Signal et Automatique de Lille, F-59000 
Lille, France}
\address[univlilleimt]{Univ. Lille, CNRS, Centrale Lille, IMT Lille Douai, UMR 9189 - CRIStAL - Centre de Recherche en Informatique Signal et Automatique
 de Lille, F-59000 Lille, France}

\begin{document}
\begin{abstract}
Spiking neural networks (SNNs) equipped with latency coding and spike-timing dependent plasticity rules offer an alternative to solve the data and energy bottlenecks of standard computer vision approaches: they can learn visual features without supervision and can be implemented by ultra-low power hardware architectures. However, their performance in image classification has never been evaluated on recent image datasets. In this paper, we compare SNNs to auto-encoders on three visual recognition datasets, and extend the use of SNNs to color images. The analysis of the results helps us identify some bottlenecks of SNNs: the limits of on-center/off-center coding, especially for color images, and the ineffectiveness of current inhibition mechanisms. These issues should be addressed to build effective SNNs for image recognition.
\end{abstract}


\maketitle

\section{Introduction}
Machine learning algorithms require good data representations to be effective~\cite{bengio12a}. Good data representations can capture underlying correlations of the data, provide invariance properties and help disentangle the data to make it linearly separable. In computer vision, much effort has been put historically into engineering the right visual features for recognizing, organizing and interpreting visual contents~\cite{lowe04a,miko05a}. More recently, and especially since the rise of deep learning, those features tend to be learned by algorithms rather than designed by human effort. Learned features have shown their superiority on a number of tasks, such as image classification~\cite{xie2017a}, image segmentation~\cite{liu15a}, and action recognition~\cite{tu18a}. Although effective, feature learning has two major drawbacks:
\begin{itemize}
	\item it is \textit{data-consuming}, as supervised learning algorithms -- especially deep learning ones -- require large amounts of annotated data to be trained;
    \item it is \textit{energy-consuming}, as training large models, e.g., using gradient descent-based algorithms, has a high computational cost, which increases with the amount of training data. These algorithms are usually run on dedicated hardware (typically GPU) that are power-intensive.
\end{itemize}

The first issue \marius{-- data consumption --} can be mitigated by the use of unsupervised learning models. Unsupervised representation learning is recognized as one of the major challenges in machine learning~\cite{bengio12a} and is receiving a growing interest in computer vision applications~\cite{coates11a,wang16a,yuan16a}. A number of unsupervised models have been developed through the years, notably auto-encoders~\cite{bourlard88a} and restricted Boltzmann machines (RBMs)~\cite{smolensky86a}, and their multi-layer counterparts, stacked auto-encoders~\cite{bengio06a} and deep belief networks (DBN)~\cite{hinton06a}. Other lines of work include sparse coding~\cite{zhang17a} and the use of semi- or weakly supervised learning algorithms~\cite{tang17a}. Moreover, in the case of neural networks, initializing a deep neural network with features learned without supervision before training can yield better generalization capabilities than purely supervised training~\cite{bengio06a}.

The second issue -- energy consumption -- is addressed much less frequently in the literature, but several authors acknowledge its importance~\cite{bekkerman11a,cao15a,hubara18a} which is 
bound to grow more and more as machine learning becomes overwhelmingly present in a large range of applications: marketing, medicine, finance, education, administration, etc. Most hardware vendors have proposed dedicated machine learning processor architectures (based on GPU, FPGA, etc.) recently~\cite{shan_tang_deep-learning-processor-list_nodate}. These hardware improvements help reduce the energy consumption by a small factor (typically one order of magnitude). Reducing further the energy consumption of learning algorithms requires to define new learning models and associated ultra-low power architectures~\cite{hubara18a,james_historical_2017,esser16a}. One promising model is spiking neural networks (SNNs). In this model, artificial neurons communicate information through spikes, as natural neurons do. Initially studied in neuroscience as a model of the brain, SNNs receive constant attention in the fields of machine learning and pattern recognition, from both the theoretical~\cite{bengio15a} and the applicative~\cite{cao15a,kamaruzaman15a,dennis15a,escobar09a} perspectives. Dedicated hardware implementing this model can be very energy-efficient~\cite{james_historical_2017}. SNNs have already shown their ability to provide near-state-of-the-art results in image classification, but only when they are trained by transferring parameters from pre-trained deep neural networks~\cite{esser16a} or by variants of back-propagation~\cite{lee16a}. In terms of energy efficiency, the first option is not viable as it still requires to train a standard deep neural network, which is exactly what should be avoided; the second option is not suited either as back-propagation is a global, centralized algorithm -- the error must be propagated from the output to all units --, whereas the efficiency of SNNs lies in their ability to perform highly decentralized, parallel processing on sparse spike data. The alternative is to use bio-inspired learning rules, such as Hebbian rules. Among those, rules based on spike-timing dependent plasticity (STDP)~\cite{bi1998synaptic} have shown promising results for learning visual features; however, they have only been evaluated on datasets with limited challenges (rigid objects, limited number of object instances, uncluttered backgrounds\ldots) such as MNIST, 3D-object, ETH-80 or NORB~\cite{querlioz2011simulation,diehl2015unsupervised,kheradpisheh2016bio,kheradpisheh2016stdp,mozafari2018first}, or on two-class datasets~\cite{kheradpisheh2016stdp,mozafari2018first}. How they perform on more complex image datasets, what is the performance gap between them and standard approaches, and what needs to be done to bridge this gap is yet to be established. 

\paragraph*{Aims and scope}
In this paper, we evaluate the ability of SNNs equipped with latency coding and STDP to learn features for visual recognition on three standard datasets (CIFAR-10, CIFAR-100, and STL-10). Our goal is to identify some of the factors that prevent STDP-based SNNs to reach state-of-the-art results on actual computer vision tasks. First, we compare the performance of SNNs on grayscale and color images (Section~\ref{section:experiments:color}), then we compare them to one standard unsupervised feature learning algorithm, sparse auto-encoders (Section~\ref{section:experiments:snn_vs_ae}). The resulting models are analyzed with respect to different factors (Section~\ref{section:discussion}): input pre-processing, feature sparsity, feature coherence, and objective functions. It allows us to identify some bottlenecks that should be tackled to bridge the gap from SNNs to state-of-the-art models. In the conclusion (Section~\ref{section:conclusion}), we suggest some solutions to help address these bottlenecks.

In this work, we consider only single-layer architectures because multi-layer SNNs with unsupervised STDP are only very recent and  difficult to train, due to the loss of spiking activity across layers~\cite{falez18a,kheradpisheh2016stdp}. To our knowledge, this is the first work that evaluates features learned by unsupervised STDP-based SNNs on recent benchmarks for object recognition and on color images, making one step towards their use for actual vision applications.

\section{Unsupervised visual feature learning}
A visual feature extractor can be modeled as a function \(f: \mathbb{R}^{h \times w} \rightarrow \mathbb{R}^{n_f}\) that maps an image or image region of size \(h \times w\) to a real vector of dimension \(n_f\). It defines a dictionary of features of size \(n_f\). In the remaining of the paper, \(f\) will denote either the feature extractor function or the resulting dictionary, depending on the context. Early visual feature extractors were handcrafted to capture specific types of visual information (e.g., distributions of edges~\cite{miko05a}). Recent approaches rather rely on machine learning to produce features that better fit the data and that can be optimized towards a specific application.

A typical learning-based feature extractor can be seen as a function \(f_\theta\) whose parameters \(\theta\) are optimized towards a specific goal by a learning algorithm. The general shape of \(f_\theta\) can be specified explicitly (e.g., a linear transform~\cite{philbin10a}), or implicitly, based on the learning algorithm to be used (e.g., in a neural network, the possible shapes for \(f_\theta\) are defined by the architecture of the network). The parameters \(\theta\) are optimized by the learning algorithm by minimizing an objective function \(L\). In a supervised setting, this optimization step can be expressed as:
\begin{equation}
\theta^* = \argmax_{\theta}\ L(X, Y; \theta)
\label{equation:supervised_objective}
\end{equation}
where \(\theta^*\) are the parameters returned by the learning algorithm, \(X=(x_1, x_2\ldots)\) denotes the training samples and \(Y = (y_1, y_2\ldots)\) the ground truth attached to the samples. \(L\) is directly set as a performance measure for the specific task to be solved, e.g. the multinomial logistic regression objective for image classification~\cite{xie2017a}, or keypoint matching for image retrieval~\cite{philbin10a}.

In an unsupervised setting, labels are not available. The optimization problem becomes:
\begin{equation}
\theta^* = \argmax_{\theta}\ L(X;\theta)
\label{equation:unsupervised_objective}
\end{equation}
In this case, \(L\) cannot be formulated towards a specific application. Some surrogate objective must be defined, that is expected to produce features that will fit the problem, e.g. image reconstruction~\cite{bourlard88a}, image denoising~\cite{vincent10a}, and maximum likelihood~\cite{hinton06a}. Learning rules are sometimes defined directly without any explicit objective function, e.g. in k-means clustering~\cite{coates11a}, but also STDP~\cite{bi1998synaptic}.

Some constraints on the parameters or learning algorithm can be added to regularize the training process and reach better solutions. These constraints reflect assumptions on properties that "good" features should have, such as:
\begin{itemize}
	\item sparsity: it is often assumed that the extracted features should be sparse, i.e. only a small number of the features can be found in a single image or image region. Sparsity is especially required when the set of features is over-complete\footnote{A dictionary of features is over-complete when its dimension (number of features) is larger than the dimension of the input (size in pixels of the images or image regions processed).}, to prevent the algorithm from reaching trivial solutions. Sparsity is commonly imposed in sparse coding~\cite{zhang17a} and auto-encoders~\cite{jiang15a};
    \item coherence of features~\cite{makhzani14a}: features should be different to span the space of visual patterns with limited redundancy\footnote{Some authors~\cite{gupta17a} claim that redundancy should rather be reached to have a good representation, but with limited evidence.}. Coherence measures the possibility to reconstruct a given feature as a weighted sum of a small number of other features, i.e. whether features are locally linearly dependent; coherence should be small for the features to be effective.
\end{itemize}
In Section~\ref{section:discussion}, these properties will serve as a basis for the analysis of the feature extractors.

\section{STDP-based feature learning}

While traditional machine learning algorithms represent data as real values, some models inspired by biology use spikes to carry information. These models are called spiking neural networks (SNNs). Working with spikes allows a complete desynchronization of the model since each spike is processed independently of the others. Thus, the system can be more energy-efficient when implemented on dedicated hardware~\cite{krichmar2015large} since no master synchronization system is required~\cite{han2016energy}. Moreover, in these models, memorization and computation are performed locally by neurons, which saves the data exchange cost and, as opposed to traditional Von-Neumann architectures, bypasses the bus bottleneck. 

In the following, we detail the preprocessing and learning mechanisms involved in SNN architectures. SNNs have two main components: the spiking neurons, discussed in Section~\ref{ssec:spiking_neuron_models}, and the synapses, which connect the neurons of the network. The synapses are generally responsible for the training of the network. One of the most used learning rule, spike-timing dependent plasticity (STDP)~\cite{bi1998synaptic}, is discussed in Section~\ref{ssec:stdp}. As SNNs are fed with spikes, a preprocessing stage is required to transform pixel values into spike trains. Section~\ref{ssec:preprocessing} is focused on the preprocessing required by SNNs to process images. In Section~\ref{ssec:neural_coding}, we discuss the representation of data as spikes, called the neural coding. The last sections describe other mechanisms required to achieve learning in SNNs: Section~\ref{ssec:ihnibition} is focused on lateral inhibition, which sets up competition between neurons to force them to learn different patterns; Section~\ref{ssec:homeostasis} presents homeostasis, which prevents certain neurons from taking advantage over the others, enforcing effective spiking activities throughout the network.

\subsection{Spiking neuron models}
\label{ssec:spiking_neuron_models}

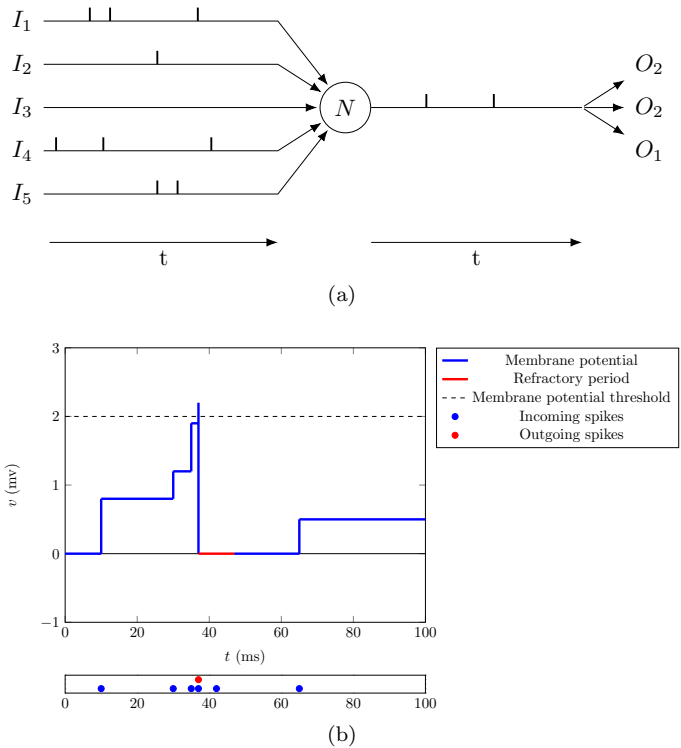
\begin{figure}[!ht]
	\centering
	\subfloat[]{
    	
    	\resizebox{.75\textwidth}{!}{\input{tikz/spiking_neuron.tikz}}
        \label{fig:spiking_neuron}
    }
    
    \subfloat[]{
    	\resizebox{.75\textwidth}{!}{\input{tikz/if.tikz}}
        \label{fig:membrane_neuron}
    }
    \caption{\ref{fig:spiking_neuron} A spiking neuron receives spike trains from incoming synapses, and generates spikes towards outgoing synapses. \ref{fig:membrane_neuron} Evolution of the membrane potential of an integrate-and-fire neuron.}
    \label{fig:if}
\end{figure}

Spiking neurons (in Figure~\ref{fig:spiking_neuron}) are defined as processing units receiving spikes from their input connections (or synapses), and emitting spikes towards their output synapses when specific input patterns are received. There are several spiking neuron models, designed to accomplish different objectives~\cite{izhikevich2004model}. Some aim to faithfully reproduce the behavior of biological neurons (e.g., the Hodgkin–Huxley and Morris–Lecar models), while others are designed to be efficiently implemented on hardware (e.g., the integrate-and-fire and Izhikevich models); the latter are usually simpler. This paper focuses on the integrate-and-fire (IF) model, to facilitate the understanding of the mechanisms of the models and to optimize the simulation speed. Most work on SNN-based image recognition use this kind of model~\cite{querlioz2011simulation,diehl2015unsupervised,kheradpisheh2016stdp}.
IF neurons have a single internal variable: their voltage $V$. When a spike is fed to a neuron, it is integrated to $V$. $V$ remains constant until the next spike, {i.e.} no leakage is applied. When $V$ reaches the neuron threshold $V_{\mathrm{th}}$, the neuron fires: it emits a spike towards its output neurons and resets its potential $V$ to $V_{\mathrm{rest}}$ (see Figure~\ref{fig:membrane_neuron}). Formally,
\begin{equation}
	V' = I, V \leftarrow V_{\mathrm{rest}} \text{ if } V \geq V_{\mathrm{th}}
\end{equation}
with $V'$ the derivative of $V$ and $I$ the input current. $V$ is also reset to $V_{\mathrm{rest}}$ between each input sample, to reset the network activity. This model allows the system to perform computations locally, since incoming spikes act on the state of the neuron, but not on the rest of the network (see Figure~\ref{fig:spiking_neuron}).

\subsection{Spike-timing dependent plasticity}
\label{ssec:stdp}

Neurons are connected via synapses, which modulate the intensity of the spikes that they transmit via their synaptic weights $W$. Adapting these weights allows long-term learning, by reinforcing or weakening connections. To allow energy-efficient hardware implementations, this learning mechanism should use only local information like the input and the output spike trains of the neuron. Finally, each synapse propagates the spikes with a random delay $d$ to introduce some noise and make lateral inhibition more active. In this paper, \(d\) is sampled from a uniform distribution in \([d_{\mathrm{min}} ; d_{\mathrm{max}}]\).

\begin{figure}[ht]
  \begin{center}
	\subfloat[Biological STDP]{\resizebox{0.45\textwidth}{!}{\input{tikz/bio_stdp.tikz}}\label{fig:bio_stdp}}
  	\subfloat[Simplified STDP]{\resizebox{0.43\textwidth}{!}{\input{tikz/simple_stdp.tikz}}\label{fig:simple_stdp}}
  \end{center}
  \caption{Weight variation w.r.t. the difference in spike timestamps.}
\end{figure}
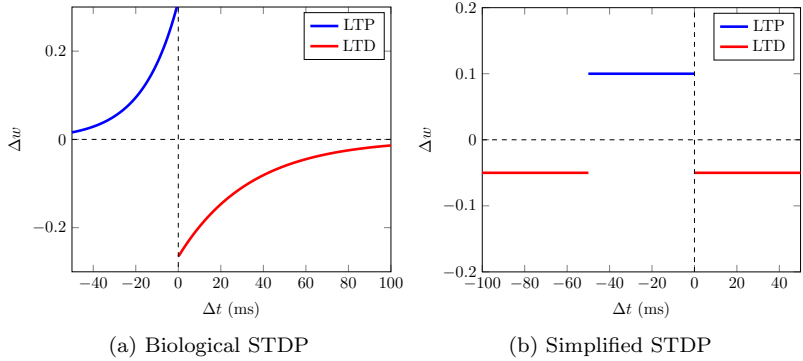

Spike-timing dependent plasticity (STDP, see Figure~\ref{fig:bio_stdp})~\cite{bi1998synaptic} is a Hebb-like learning rule observed in biology. It reinforces the connections between neurons that have correlated firing patterns (potentiation), and weakens the others (depression). Multiplicative STDP~\cite{querlioz2011simulation} (see Figure~\ref{fig:simple_stdp}) is used in this paper. When a neuron fires a spike at time $t_{\mathrm{post}}$, all incoming synapses are updated as follows:
\begin{equation}
	\Delta w = 
    \left \{
   \begin{array}{r l}
     \alpha_+ e^{-\beta_+\frac{w-W_{\mathrm{min}}}{W_{\mathrm{max}}-W_{\mathrm{min}}}} & \text{if } t_{\mathrm{post}} \geq t_{\mathrm{pre}} \\
     -\alpha_- e^{-\beta_-\frac{W_{\mathrm{max}}-w}{W_{\mathrm{max}}-W_{\mathrm{min}}}} & \text{otherwise}
   \end{array}
   \right .
\end{equation}
with $\Delta_w$ the update applied to synapse weight \(w\), $t_{\mathrm{pre}}$ the timestamp of the input spike and $t_{\mathrm{post}}$ the timestamp of the output spike. $\alpha_+$ and $\alpha_-$ are respectively the learning rates applied for potentiation and depression. $\beta+$ and $\beta-$ control the slope of the exponential. $W_{\mathrm{min}}$ and $W_{\mathrm{max}}$ are the bounds for synaptic weights. When no spike occurs during the presentation of a sample, $t_{\mathrm{pre}}$ is set to $+\infty$, so that weight depression occurs.

\subsection{Image pre-processing and color handling}
\label{ssec:preprocessing}
As STDP learns correlations between input spikes, images are usually pre-processed to help STDP find meaningful correlations. Typically, edges are extracted from the grayscale images, e.g. through a difference-of-Gaussian (DoG) filter~\cite{kheradpisheh2016stdp,tavanaei18a} or Gabor filters~\cite{kheradpisheh2016bio}. In this paper, we use on-center/off-center coding, which combines DoG filtering with a 2-channel representation necessary to encode the data as spikes. This coding is inspired from bipolar cells of the retina. The output of the DoG filter at position \((x, y)\) of image \(I\) is defined as:
\begin{align}
	&\mathrm{DoG}(x, y) = I(x,y) \ast (G_{\mathrm{DoG}_{\mathrm{size}}, \mathrm{DoG}_{\mathrm{center}}} - G_{\mathrm{DoG}_{\mathrm{size}}, \mathrm{DoG}_{\mathrm{surround}}}) \nonumber
\end{align}
where \(\ast\) is the convolution operator and \(G_{S, \sigma}\) is a normalized Gaussian kernel of size \(S\) and scale \(\sigma\) defined as:
\begin{align}
	&G_{S, \sigma}(u,v) =\frac{g_\sigma(u, v)}{\sum\limits_{i = -\mu}^\mu \sum\limits_{j = -\mu}^\mu g_\sigma(i, j)}, u, v \in [-\mu, \mu], \mu = \frac{S}{2}, \nonumber
\end{align}
with $g_\sigma$ the centered 2D Gaussian function of variance \(\sigma\). The parameters of the filter are its size $\mathrm{DoG}_{\mathrm{size}}$ and the variances of the Gaussian kernels $\mathrm{DoG}_{\mathrm{center}}$ and $\mathrm{DoG}_{\mathrm{surround}}$. After DoG filtering, the image is split into two channels \(c_+\) and \(c_-\) as follows:
\begin{align*}
     c_\star(x,y) &= \max(0, \star{}\mathrm{DoG}(x,y)), \star \in \{+, -\}
\end{align*}
where \(\mathrm{DoG}(x,y)\) denotes the output of the DoG filter at position \((x,y)\). \(c_+\) is the positive ("on") channel, and \(c_-\) the negative ("off") channel. This coding makes it possible to encode negative DoG values as spikes. No thresholds are applied to \(c_+\) and \(c_-\), i.e. the output values of the DoG filter are not filtered. 

Color processing in SNNs is a problem that has not been addressed much in the literature so far. 
To apply on-center/off-center coding to color images, we define two strategies. In the first strategy, called RGB color opponent channels, the coding is applied to channels computed as differences of pairs of RGB channels: red-green, green-blue, and blue-red. The second strategy is inspired by biological observations: in the lateral geniculate nucleus, which mainly connects the retina to the visual cortex, three types of color channels exist: the black-white opponent channel (which corresponds to the grayscale image), the red-green opponent channel, and the yellow-blue opponent channel~\cite{livingstone1984anatomy}. The second strategy applies on-center/off-center coding to the red-green and yellow-blue (computed as $0.5\times R+0.5 \times G-B$) channels. This leads to four possible configurations of image coding: grayscale only, RGB opponent channels, biological color opponent channels (referred to as Bio-color), and the combination of the grayscale channel and the Bio-color channels.

\subsection{Neural coding}
\label{ssec:neural_coding}
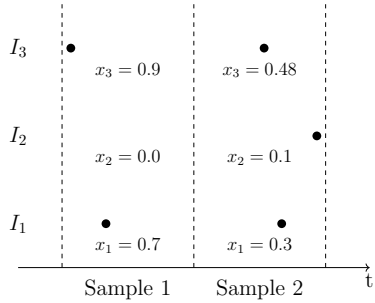
\begin{figure}[ht]
\begin{center}
	\resizebox{!}{4cm}{\input{tikz/coding.tikz}}
\end{center}
\caption{Latency coding.}
\label{fig:coding}
\end{figure}

A step called neural coding is required to transform real or integer values from datasets spike trains. Several neural coding \marius{techniques} have been proposed in the literature. The most common ones are frequency coding, which encodes values directly as frequencies of spike trains, and temporal coding, which encodes values as timestamps of single spikes~\cite{brette2015philosophy} (see Figure~\ref{fig:coding}). This paper uses latency coding, a form of temporal coding, since it has several advantages over frequency coding in visual tasks~\cite{rullen2001rate}. It allows to represent a sample with fewer spikes, but also to simplify the model since at most one spike per connection is emitted during the presentation of a sample. Each input value $x$ is transformed into a spike timestamp $t$ as follows:
\begin{equation}
	t = (1.0-x)\times T_{\mathrm{duration}}
\end{equation}
with $x\in[0, 1]$ the input value and $T_{\mathrm{duration}}$ the duration of the presentation of a data sample.

\subsection{Inhibition}
\label{ssec:ihnibition}

STDP, or more generally unsupervised and local learning rules, require some competition mechanisms in order to prevent all neurons from learning the same pattern. Lateral inhibition is a straightforward method to do so in SNNs~\cite{querlioz2011simulation}: when a neuron fires, it sends inhibitory spikes to others neurons of the same layer, which reduce their abilities to fire. Only a small number of neurons is active at each input, which makes it possible to learn different patterns. In this paper, winner-take-all (WTA) competition is used: only one neuron is allowed to spike at each position per sample.

\subsection{Homeostasis}
\label{ssec:homeostasis}

\begin{figure}[ht]
	\begin{center}
	\subfloat[$t_{\mathrm{obj}} = 0.5$]{
    	\resizebox{!}{.6cm}{
    	\begin{tabular}{cccc}
    		\fbox{\includegraphics{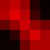}} &
    		\fbox{\includegraphics{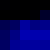}} &
   			\fbox{\includegraphics{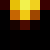}} &
    		\fbox{\includegraphics{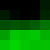}} \\
    	\end{tabular}
        \label{fig:f_ex_0d5}
        }
    }
	\subfloat[$t_{\mathrm{obj}} = 0.7$]{
    	\resizebox{!}{.6cm}{
        \begin{tabular}{cccc}
          \fbox{\includegraphics{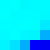}}&
          \fbox{\includegraphics{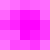}}&
          \fbox{\includegraphics{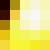}}&
          \fbox{\includegraphics{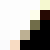}}\\
    	\end{tabular}
        \label{fig:f_ex_0d7}
        }
    }
	\end{center}
    \caption{Filters learned with different $t_{\mathrm{obj}}$.}
    \label{fig:f_ex}
\end{figure}

Several authors demonstrated experimentally than SNNs need homeostasis to guarantee an effective learning process~\cite{querlioz2011simulation,diehl2015unsupervised}. It can be done by adapting the thresholds $V_{\mathrm{th}}$ to prevent one neuron from dominating the others. The thresholds of neurons that fire often are increased, so they will tend to fire less often later on; inversely, the thresholds of sub-active neurons are decreased. Since latency coding is used, we train neurons to fire at an objective timestamp $t_{\mathrm{obj}}$. The choice of this parameter can be tricky. An early $t_{\mathrm{obj}}$ trains neurons to respond to few input spikes, so that neurons learn only local patterns in few input channels (blobs or edges -- see Figure~\ref{fig:f_ex_0d5}). A late $t_{\mathrm{obj}}$ results in larger patterns with a combination of multiple input channels (see Figure~\ref{fig:f_ex_0d7}).

Threshold adaptation is performed as follows:
\begin{equation}
	\Delta {V_{\mathrm{th}}} = -\eta\times(t_{\mathrm{fire}}-t_{\mathrm{obj}})+
     \left \{
   \begin{array}{c l}
     \eta & \text{if } \text{the neuron is the first to fire} \\
     -\frac{\eta}{N_{\mathrm{neuron}}-1} & \text{otherwise}
   \end{array}
   \right .
\end{equation}
with $\eta$ the learning rate, $t_{\mathrm{fire}}$ the timestamp of the spike fired by the neuron, $t_{\mathrm{obj}}$ the objective timestamp and $N_{\mathrm{neuron}}$ the number of output neurons. Setting large initial values for the thresholds may prevent the neurons from firing. In the absence of neuronal activity, no learning nor threshold adaptation can be performed. It is therefore preferable to initialize the thresholds with small values to promote neuronal activity within the network.

\subsection{Spike to feature conversion}

Finally, spikes need to be transformed back into feature values that will be fed to the classifier. Feature values are computed as follows:
\begin{equation}
	f_i = 1.0-\frac{t-T_{\mathrm{output}_{\mathrm{min}}}}{T_{\mathrm{output}_{\mathrm{max}}}-T_{\mathrm{output}_{\mathrm{min}}}}
\end{equation}
with $f_i$ the \(i\)-th output feature value, $t$ the spike timestamp (if no spike occurs, then $t$ is set to $T_{\mathrm{output}_{\mathrm{max}}}$), $[T_{\mathrm{output}_{\mathrm{min}}}$, $T_{\mathrm{output}_{\mathrm{max}}}]$ the range of possible timestamps. $T_{\mathrm{output}_{\mathrm{min}}}$ can be computed as $T_{input_{min}}+d_{\mathrm{min}}$, with $d_{\mathrm{min}}$, and $T_{\mathrm{output}_{\mathrm{max}}}$ as $T_{input_{max}}+d_{\mathrm{max}}$.

\section{Learning visual features with sparse auto-encoders}
\label{section:auto-encoders}
Auto-encoders (AEs)~\cite{bourlard88a} are unsupervised neural networks that learn latent representations that allow to best reconstruct the input data. In this work, among all unsupervised feature learning algorithms, we only consider single-layer AEs, for two reasons. First, they belong to the family of neural networks, as SNNs do, and, within this family, they are one of the most representative models for unsupervised learning (its main competitor being RBMs, which have been shown to optimize a similar criterion~\cite{vincent10a} and yield comparable performance for visual feature learning~\cite{coates11a}). Then, we restrict our approach to single-layer networks, as multi-layer SNNs are only starting to emerge~\cite{kheradpisheh2016stdp}; we believe one-layer SNNs should be well mastered before addressing multi-layer architectures.

The typical architecture of an AE is organized in two parts: an encoder \(\mathrm{enc}\) mapping the input \(X\) to its latent representation \(Z = \mathrm{enc}(X)\), and a decoder \(\mathrm{dec}\) computing a reconstruction \(\tilde{X}\) of the input from its latent representation: \(\tilde{X} = \mathrm{dec}(Z) = \mathrm{dec}(\mathrm{enc}(X))\).
The objective function (Eq.~\ref{equation:unsupervised_objective}) thus becomes:
\begin{equation}
\theta^* = \argmin_{\theta}\ L(X,\tilde{X};\theta)
\end{equation}
where \(L(.,.;\theta)\) is some measure of the dissimilarity between the input \(X\) and its reconstruction \(\tilde{X}\) given the model \marius{parameterized} by \(\theta\); in other words, the auto-encoder aims at reconstructing its input with minimal reconstruction error. \marius{In our experiments, the Euclidean distance measures the reconstruction error}. 

The encoder and the decoder can be defined as single-layer or multilayer (in the case of stacked AEs) neural networks. In the following, we will consider only single-layer models of this form:
\begin{equation}
\begin{array}{c}
Z = \mathrm{enc}(X) = \sigma(W_\mathrm{enc}X + b_\mathrm{enc})\\
\mathrm{dec}(Z) = W_\mathrm{dec}Z + b_\mathrm{dec}\\
\end{array}
\label{equation:single_layer_autoencoder}
\end{equation}
where \(W_\mathrm{enc} \in \mathbb{R}^{h\times w, n_f}\) (resp. \(W_\mathrm{dec} \in \mathbb{R}^{n_f, h\times w}\)) is the weight matrix of the connections in the encoder (resp. the decoder), \(b_\mathrm{enc} \in \mathbb{R}^{n_f}\) (resp. \(b_\mathrm{dec} \in \mathbb{R}^{h\times w}\)) is the bias vector of the encoder (resp. the decoder), and \(\sigma(.)\) is some activation function\footnote{We only consider models where no activation function is applied to the decoder output as we work on continuous (image) data in \([0,1]\).}, in our case the sigmoid activation function \(\sigma(x) = \frac{1}{1+e^{-x}}\). The output of the encoder correspond to the visual features learned by the AE: \(f = \mathrm{enc}(X)\).

To make the AE learn useful representations, the initial approach was to impose an information bottleneck on the model, by learning representations with dimensionalities lower than the ones of the input data (\(n_f < h \times w\)). However, such  representations cannot capture the richness of the visual information, so current approaches rather use over-complete (\(n_f > h \times w\)) representations. In this case, some additional constraints must be enforced on the model to prevent it from learning trivial solutions, e.g., the identity function. These constraints generally take the form of an additional term in the objective function, for instance: weight regularization, explicit sparsity constraints (sparse AEs~\cite{coates11a,jiang15a}, k-sparse AEs~\cite{makhzani14a}) or regularization of the Jacobian of the encoder output \(Z\) (contractive AEs~\cite{rifai11a}). Another approach is to change the objective function from reconstruction to another criterion, for instance data denoising~\cite{vincent10a}.

In this paper, we will consider sparse AEs as a baseline to assess the performance of STDP-based feature learning. More recent models (denoising AEs~\cite{vincent10a}, contractive AEs~\cite{rifai11a}, etc.) can reach better performance, but sparse AEs are closer to current STDP-based SNNs, which also feature explicit sparsity constraints, usually through lateral inhibition. Also, it allows us to set a minimum bound that SNNs should at least reach to be competitive with regular feature learning algorithms, and identify some directions to follow to achieve this goal; it constitutes a first step before taking STDP-based SNNs further.

In the following, we describe the weight regularization and sparsity constraint terms that we used in our experiments~:
\begin{itemize}
	\item L2 weight regularization: \(\frac{\lambda}{2}(||W_\mathrm{enc}||_2^2 + ||W_\mathrm{dec}||_2^2)\), where \(||.||_2\) denotes the Frobenius norm and \(\lambda\) is the weight decay parameter;
 	\item sparsity term~\cite{jiang15a}: \(\gamma.\mathrm{KL}(\hat{\rho}||\rho)\), where \(\rho\) is the desired sparsity level of the system, \(\hat{\rho}\) is the vector of average activation values of the hidden neurons over a batch, \(\mathrm{KL}(.||.)\) the Kullback-Liebler divergence, and \(\gamma\) the weight applied to the sparsity term in the objective function.
\end{itemize}
This yields the final objective function for the AE:
\begin{equation}
L(X,\tilde{X};\theta) = \frac{1}{2}||X - \tilde{X}||^2_2 + \frac{\lambda}{2}(||W_\mathrm{enc}||_2^2 + ||W_\mathrm{dec}||_2^2) + \gamma.\mathrm{KL}(\hat{\rho}||\rho)
\label{equation:autoencoder:objectove_function_complete}
\end{equation}

\section{Experiments}
\label{section:experiments}
\subsection{Experimental protocol}
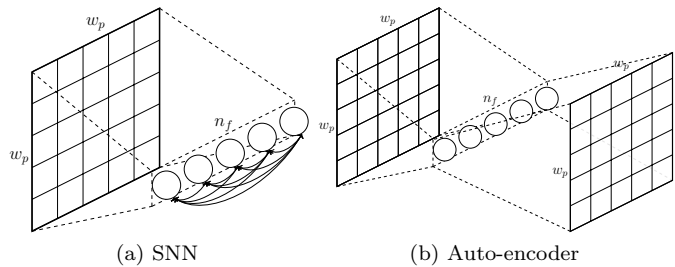
\begin{figure}[ht]
\begin{center}
	\subfloat[SNN]{\resizebox{!}{3cm}{\input{tikz/topology.tikz}}}
    \subfloat[Auto-encoder]{\resizebox{!}{3cm}{\input{tikz/ae.tikz}}}
\end{center}
\caption{(a) SNN architecture used in the experiments. Solid arrows denote inhibitory connections between hidden units. (b) AE architecture used in the experiments.}
\label{figure:snn_architecture}
\end{figure}

The SNN and AE architectures used in our experiments are single-layer networks with \(n_f\) hidden units (see Figure~\ref{figure:snn_architecture}). We follow the experimental protocol proposed by Coates~\textit{et al.}~\cite{coates11a} to compare unsupervised feature extractors. It is organized in two stages, described below: visual feature learning, and the evaluation of the learned features on image classification benchmarks.

\paragraph{Feature learning}
From the training image dataset \(\mathcal{I} = (I_1, I_2, \ldots, I_n)\), we randomly sample \(n_p\) patches of size \(w_p \times w_p\). The patches are fed to the feature learning algorithm for training, to produce a dictionary of \(n_f\) features.

\begin{figure}[ht]
\begin{center}
	\resizebox{0.8\textwidth}{!}{\input{tikz/classification.tikz}}
\end{center}
\caption{Experimental protocol. (a) Input image, where $k \times k$ patches of size $w_p \times w_p$ are extracted with a stride $s$. (b) \(n_f\) feature maps of size $k \times k$ produced by the feature extractor from its dictionary of $n_f$ features. (c) Output vector constructed by sum pooling over $r \times r$ regions of the feature maps.}
\label{figure:experimental_protocol}
\end{figure}
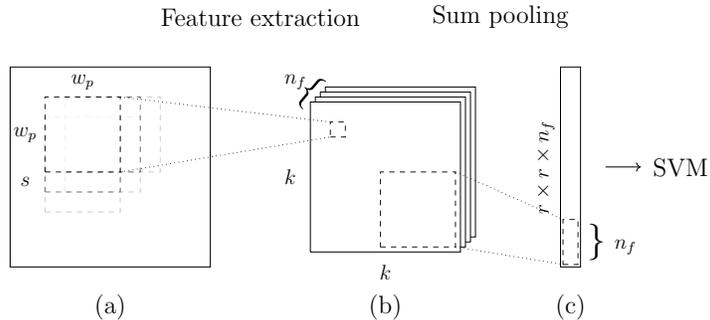

\paragraph{Image recognition}
The learned feature dictionary is used to produce image descriptors that are fed to a classifier following this process (Figure~\ref{figure:experimental_protocol}):
\begin{enumerate}
	\item Image patches of size \(w_p \times w_p\) are densely sampled from the images with stride \(s\), producing \(k \times k\) patches per image (Figure~\ref{figure:experimental_protocol}a).
    \item Patches are fed to the feature extractor, producing \(k \times k\) feature vectors of dimension \(n_f\) per image, organized into feature maps where each position corresponds to one patch of the input image (Figure~\ref{figure:experimental_protocol}b).
    \item We apply sum pooling over a grid of size \(r \times r\): the feature vectors of the patches within each grid cell are summed to produce a unique vector of size \(n_f\) per cell. These vectors are then concatenated to produce a single feature vector of size \(r \times r \times n_f\) for each image (Figure~\ref{figure:experimental_protocol}c).
    \item The feature vectors of the images are fed to a linear support vector machine (SVM) for training (training set) or classification (test set).
\end{enumerate}

\subsection{Datasets}
We perform experiments on three datasets commonly used to evaluate unsupervised feature learning algorithms: CIFAR-10, CIFAR-100, and STL-10. Table~\ref{table:datasets} provides the properties of these datasets. Since previous work evaluated SNNs only on grayscale images, we also use grayscale versions of the three datasets, referred to as CIFAR-10-bw, CIFAR-100-bw, and STL-10-bw.

\begin{table}[ht]
  \centering
  \begin{tabular}{|c|c|c|c|c|}
    \hline
    Dataset & Resolution & \# classes & Training set size & Test set size\\
    \hline
    CIFAR-10~\cite{krizhevsky09a} & 32 \(\times\) 32 & 10 & 50,000 & 10,000\\
    \hline
    CIFAR-100~\cite{krizhevsky09a} & 32 \(\times\) 32 & 100 & 50,000 & 10,000\\
    \hline
    STL-10~\cite{coates11a} & 96 \(\times\) 96 & 10 & 5,000 & 8,000\\
    \hline
  \end{tabular}
  \caption{Properties of the datasets used in the experiments.}
  \label{table:datasets}
\end{table}

Contrary to MNIST, which is the preferred dataset in the SNN literature~\cite{tavanaei18a}, these datasets provide color images of actual objects rather than just binary images of digits. It makes it possible to evaluate SNNs in more realistic conditions, in terms of data richness and importance of image pre-processing. Also, unlike MNIST, but also other datasets such as NORB, they are not solved or nearly-solved problems (classification accuracy above 95\%), so the results can highlight better the properties of the algorithms.

\subsection{Implementation details}
We use image patches of size \(5\times 5\) pixels (\(w_p=5\)) and a stride \(s=1\). We evaluate the algorithms with two sizes of feature dictionaries, \(n_f=64\) and \(n_f=1024\). To produce final image descriptors, features are pooled over \(2 \times 2\) image regions (\(r=2\)), yielding image descriptors of size \(4 \times n_f\).

We used a grid search to find the optimal parameters for the AEs and we only report results for the best configuration for each experimental setting. Table~\ref{table:experiments:AE_parameters} provides the values of the parameters that we retained. They were consistently optimal over datasets. The AEs are trained for 1,000 epochs on 200,000 random patches from the training set considered. We use the Adadelta optimizer~\cite{zeiler12a} with an initial learning rate \(\mathrm{lr} = 1.0\). AEs are implemented using TensorFlow.

\begin{table}[t]
\centering
\begin{tabular}{|c|c|c|c|c|}
\hline
Data & \(n_f\) & \(\rho\) & \(\gamma\) & \(\lambda\)\\
\hline
\multirow{2}{*}{Color} & 64 & 0.005 & 0.5 & \(10^{-4}\)\\
\cline{2-5}
 & 1024 & 0.005 & 0.1 & \(10^{-5}\)\\
\hline
\multirow{2}{*}{Grayscale} & 64 & 0.01 & 0.05 & \(10^{-5}\)\\
\cline{2-5}
& 1024 & 0.005 & 0.1 & \(10^{-5}\)\\
\hline
\end{tabular}
\caption{AE parameters used in the experiments.}
\label{table:experiments:AE_parameters}
\end{table}

Table~\ref{table:parameters} provides the parameters used to train SNNs. These parameters values were obtained using a greedy search: the optimal value of each parameter was searched while the values of all other parameters were fixed. The large number of parameters in this model did not allow us to perform a full grid search on all parameters. All SNN models are trained on 100,000 random patches from the training sets for 100 epochs.
 
 \begin{table}[t]
\begin{center}
\begin{tabular}{|lr|lr|}
    \hline
    \multicolumn{4}{|c|}{Neuron}\\
    \hline
    $V_{\mathrm{th}}(0)$ & 20 mv & $V_{\mathrm{rest}}$ & 0 mv\\
    \hline
    \multicolumn{4}{|c|}{STDP}\\
    \hline
    $W_{\mathrm{min}}$ & 0.0 & $W_{\mathrm{max}}$ & 1.0\\
    $d_{\mathrm{min}}$ & 0.0 & $d_{\mathrm{max}}$ & 0.01\\
    $\alpha_+$ & 0.001 & $\alpha_-$ & 0.001\\
    $\beta_+$ & 1.0 & $\beta_-$ & 1.0\\
    \hline
\end{tabular}
\begin{tabular}{|lr|lr|}
    \hline
    \multicolumn{4}{|c|}{Neural Coding}\\
    \hline
    $T_{\mathrm{duration}}$ & 1.0 & &\\
    \hline
    \multicolumn{4}{|c|}{Threshold Adaptation}\\
    \hline
	$t_{\mathrm{obj}}$ & 0.7 & $\eta$ & 0.001\\
    \hline
    \multicolumn{4}{|c|}{Pre-processing}\\
    \hline
    $\mathrm{DoG}_{\mathrm{center}}$ & 1.0 & $\mathrm{DoG}_{\mathrm{surround}}$ & 2.0\\
    $\mathrm{DoG}_{\mathrm{size}}$ & 7 & &\\
    \hline
\end{tabular}
\end{center}
\caption{SNN parameters used in the experiments.}
\label{table:parameters}
\end{table}

Classification was performed using LibSVM~\cite{chang01a} with a linear kernel and default parameters. All reported accuracies are averaged over three runs of the feature learning algorithms.

\subsection{Color processing with SNNs}
\label{section:experiments:color}
\begin{figure}[ht]
	\begin{center}
	\resizebox{0.6\textwidth}{!}{\includegraphics{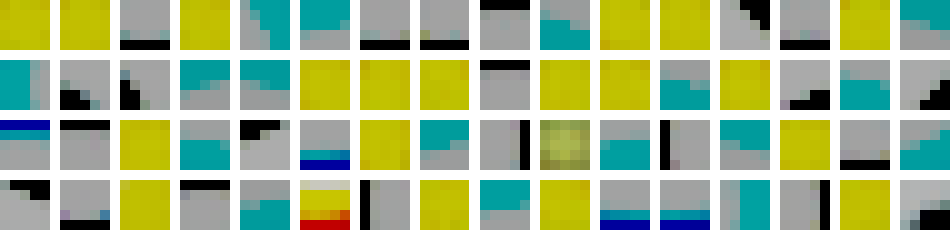}}
	\end{center}
    \caption{Examples of SNN features learned on raw RGB pixels (trained on CIFAR-10). They are mostly dead units or simple repeated patterns.}
    \label{fig:raw_snn}
\end{figure}

We first evaluate the strategies to encode color discussed in Section~\ref{ssec:preprocessing}: images are first encoded using one of these strategies, then on-center/off-center coding is applied to each channel. Table~\ref{tab:snn_color_results} shows the classification accuracies yielded by each strategy on every dataset. Both color coding techniques, biological channels and RGB opponent channels, provide similar accuracies. However, using grayscale images yields better results than color images. This is counter-intuitive, since color images contain all the information available from grayscale images. Since the SNN processes all inputs in the same way, on-center/off-center coding must cause this information loss. However, this preprocessing step is currently required to extract edges from the images and feed the SNN inputs with spike trains that represent specific visual information. Training an SNN directly from RGB images could be an alternative, but is very challenging, because the active pixels in the patterns to learn can vary (from zero for a black pattern to the size of a patch when all pixels have the maximum intensity); it cannot be handled by existing homeostasis models. Figure~\ref{fig:raw_snn} shows examples of filters learned from raw RGB images; since the network has a single layer, the filter image for one neuron is obtained by simply interpreting the normalized weights of its input synapses as RGB values. Many filters converge towards similar or uninformative patterns. It results mostly in dead units and repeated features.

\begin{table}[ht]
\centering
\begin{tabular}{|c|c|c|c|}
\hline
Dataset & Color coding & \(n_f=64\) & \(n_f=1024\)\\
\hline
\multirow{4}{*}{CIFAR-10} & RGB opponent & 37.66 $\pm$ 0.73  & 45.04 $\pm$ 0.06\\
\cline{2-4}
 & Bio-color & 37.53 $\pm$ 0.33 & 43.54 $\pm$ 0.07 \\
\cline{2-4}
 & Grayscale & 45.37 $\pm$ 0.13 & 52.78 $\pm$ 0.41 \\
\cline{2-4}
 & Grayscale + color & 48.27 $\pm$ 0.47 & 56.93 $\pm$ 0.59 \\
\hline
\multirow{4}{*}{CIFAR-100} & RGB opponent & 17.14 $\pm$ 0.22 & 19.87 $\pm$ 0.03\\
\cline{2-4}
 & Bio-color & 17.06 $\pm$ 0.09 & 19.19 $\pm$ 0.35\\
\cline{2-4}
 & Grayscale & 18.43 $\pm$ 0.34 & 22.67 $\pm$ 0.36 \\
\cline{2-4}
 & Grayscale + color & 25.20 $\pm$ 0.76 & 30.44 $\pm$ 0.48 \\
\hline
\multirow{4}{*}{STL-10} & RGB opponent& 44.13 $\pm$ 1.30 & 51.20 $\pm$ 0.30\\
\cline{2-4}
 & Bio-color & 44.23 $\pm$ 0.41 & 50.95 $\pm$ 0.08 \\
\cline{2-4}
 & Grayscale &  44.66 $\pm$ 0.87 & 51.40 $\pm$ 0.69 \\
\cline{2-4}
 & Grayscale + color & 49.20 $\pm$ 1.04 & 54.34 $\pm$ 0.30 \\
\hline
\end{tabular}
\caption{Classification accuracy (\%) w.r.t. to the color coding strategy.}
\label{tab:snn_color_results}
\end{table}

Finally, we evaluate the combination of color and grayscale images by training half of the features on each input independently. Results in Table~\ref{tab:snn_color_results} show that it performs best, showing that DoG-filtered color images still contain information that grayscale DoG-filtered images do not contain. In the remaining of the paper, all the runs performed on color images use this strategy.

\subsection{SNNs versus AEs}
\label{section:experiments:snn_vs_ae}
The classification accuracies for each feature learning algorithm and dataset are reported in Table~\ref{table:results:accuracy}. AEs perform consistently better than SNNs\footnote{Since we chose to use a simple sparse AE (see Section~\ref{section:auto-encoders}), the actual gap between SNNs and state-of-the-art models should be larger than what these experiments show.}. So, how to bridge the gap between STDP learning and standard neural network approaches? Several elements may explain the performance of STDP. The results reported in Table~\ref{table:results:accuracy} show two trends. First, working with colors always yields better results than working with grayscale images; a straightforward explanation is that color is significant to recognize objects in the datasets considered, either because natural objects (e.g., animals) represented in the datasets have a limited, meaningful set of colors, either because the contexts of the objects (e.g., the sky behind airplanes) have meaningful colors. The second trend is that the performance gap between SNNs and AEs is larger on color images than on grayscale images, showing that SNNs cannot handle color well, at least not with the straightforward color coding \marius{techniques} that were used in the experiments. This result highlights the importance of color in object recognition, and therefore the need for a more efficient neural \marius{coding} of color in SNNs.

\begin{table}[ht]
\centering
\begin{tabular}{|c|c|c|c|c|}
\hline
\multirow{2}{*}{Dataset} & \multicolumn{2}{c|}{SNN} & \multicolumn{2}{c|}{AE}\\
\cline{2-5}
& \(n_f=64\) & \(n_f=1024\) & \(n_f=64\) & \(n_f=1024\)\\
\hline
CIFAR-10 & 48.27\(\pm\)0.47 & 56.93\(\pm\)0.59 & 57.56\(\pm\)0.08 & 66.98\(\pm\)0.33\\
\hline
CIFAR-10-bw & 45.37\(\pm\)0.13 & 52.77\(\pm\)0.41 & 53.69\(\pm\)0.34 & 59.50\(\pm\)0.17\\
\hline
CIFAR-100 & 25.20\(\pm\)0.76 & 30.45\(\pm\)0.48 & 37.71\(\pm\)0.19 & 36.43\(\pm\)0.29\\
\hline
CIFAR-100-bw & 18.43\(\pm\)0.34 & 22.67\(\pm\)0.36  & 23.62\(\pm\)0.18 & 26.56\(\pm\)0.05\\
\hline
STL-10 & 49.20\(\pm\)1.04 & 54.34\(\pm\)0.30 & 52.28\(\pm\)0.47 & 55.74\(\pm\)0.25\\
\hline
STL-10-bw & 44.66\(\pm\)0.87 & 51.40\(\pm\)0.69 & 50.63\(\pm\)0.23 & 52.88\(\pm\)0.29\\
\hline
\end{tabular}
\caption{Average classification accuracy (\%) and its standard deviation w.r.t. to the datasets and feature learning algorithms.}
\label{table:results:accuracy}
\end{table}

\begin{figure}[ht]
	\begin{center}
	\subfloat[N=64, Grayscale]{
    	\resizebox{0.3\textwidth}{!}{\includegraphics{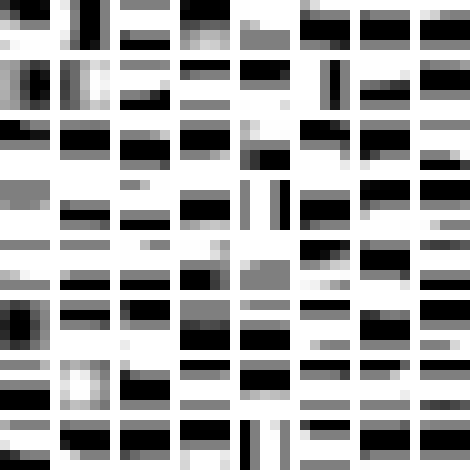}}
    }
	\subfloat[N=1024, Grayscale]{
    	\resizebox{0.3\textwidth}{!}{\includegraphics{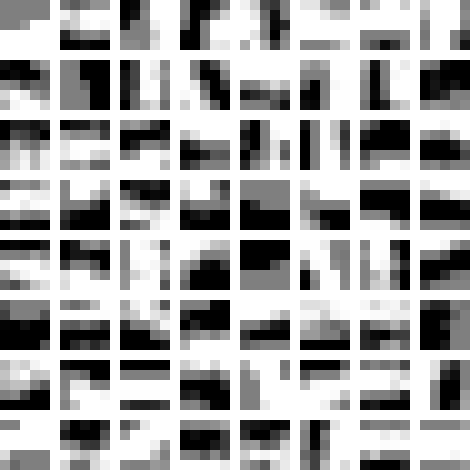}}
    }
    
   \subfloat[N=64, Color]{
    	\resizebox{0.3\textwidth}{!}{\includegraphics{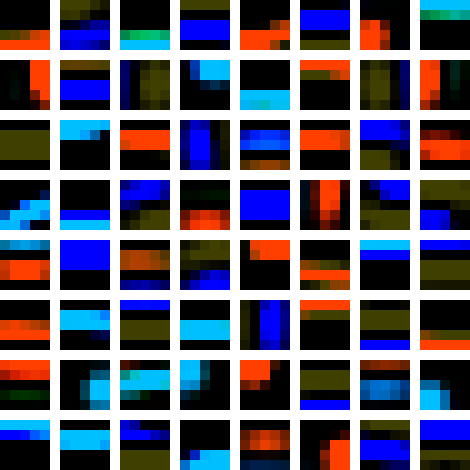}}
    }
    \subfloat[N=1024, Color]{
    	\resizebox{0.3\textwidth}{!}{\includegraphics{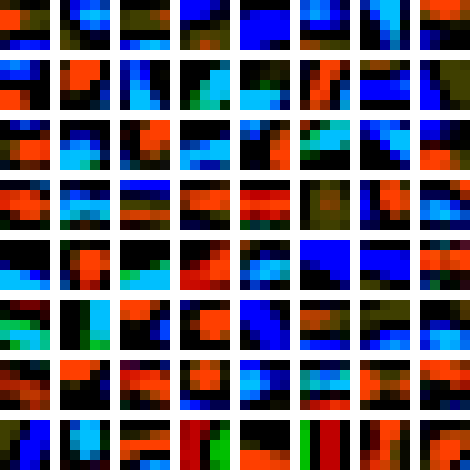}}
    }
	\end{center}
    \caption{Grayscale and color filters learned by SNNs on CIFAR-10-bw and CIFAR-10. For \(n_f=1024\), random samples are shown.}
    \label{figure:experiments:SNN_filters}
\end{figure}

\begin{figure}[ht]
	\begin{center}
	\subfloat[N=64, Grayscale]{
    	\resizebox{0.3\textwidth}{!}{\includegraphics{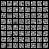}}
    }
	\subfloat[N=1024, Grayscale]{
    	\resizebox{0.3\textwidth}{!}{\includegraphics{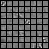}}
    }
    
   \subfloat[N=64, Color]{
    	\resizebox{0.3\textwidth}{!}{\includegraphics{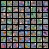}}
    }
    \subfloat[N=1024, Color]{
    	\resizebox{0.3\textwidth}{!}{\includegraphics{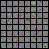}}
    }
	\end{center}
    \caption{Grayscale and color filters learned by AEs on CIFAR-10-bw and CIFAR-10. For \(n_f=1024\), random samples are shown.}
    \label{figure:experiments:AE_filters}
\end{figure}

Looking at the filters learned by SNNs and AEs provides additional information about the properties of features learned by STDP and potential reasons for the performance gap. Figures~\ref{figure:experiments:SNN_filters} and \ref{figure:experiments:AE_filters} show samples of filters learned by SNNs and AEs, respectively. The filters are different in nature. Filters learned by STDP are mostly edges, and some blobs, that are well-defined, with one or two dominant colors. By contrast, AEs learn more complex features; edges and blobs can still be observed, but they include a larger range of color or gray levels and are not as elementary as the ones learned by SNNs. Simple, well-defined features like the ones learned by STDP are conceptually pleasing because they represent elementary object shapes that can easily be understood. They suggest better generalization abilities from the feature extractor, and correspond to biological observations~\cite{kheradpisheh2016stdp}. However, they are not as effective in practice. AEs can also produce features closer to the ones obtained with SNNs (although with larger ranges of tones and intensities), but we could observe such features only by increasing the weight of L2 regularization, usually at some cost in accuracy.

\begin{figure}[ht]
  \begin{center}
  	\subfloat[$\beta$ = 1.0]{
    	\resizebox{0.4\columnwidth}{!}{\input{tikz/w_histo.tikz}}
    }
  	\subfloat[$\beta$ = 4.0]{
    	\resizebox{0.4\columnwidth}{!}{\input{tikz/w_histo2.tikz}}
    }
  \end{center}
  \caption{Distribution of weights (log. scale) in an SNN (\(n_f=64\)) after training w.r.t. $\beta$. Most weights have values close to 0 or 1 when $\beta$ decreases.}
  \label{fig:stdp_weight}
\end{figure}
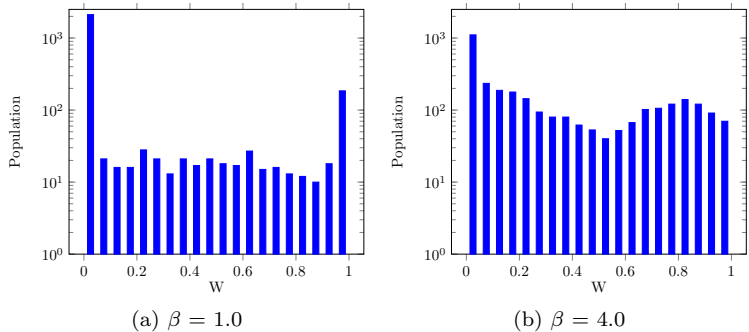

The specific looks of SNN features can be explained in two ways. First, the use of on-center/off-center coding as a preprocessing step biases the models towards edge-like filters, as it highlights the edges in the images.
Moreover, the learned features contain exclusively black or saturated colors because STDP rules tend towards a saturating regime for weights: once a given unit has learned a pattern, repeated expositions to this pattern will reinforce the sensitivity to this pattern until the weights reach either 1 or 0. This is illustrated in Figure~\ref{fig:stdp_weight}a, which shows the distribution of weights in an SNN after training: most weight values are close to 0 or 1. Since AEs perform better and have more staggered weights, one may believe that saturated weights are detrimental to the performance of SNNs. To check this, we performed experiments with different values for the parameters \(\beta_+\) and \(\beta_-\): increasing their values allow the weights to "escape" more easily from their limit values \(W_{\mathrm{min}}\) and \(W_{\mathrm{max}}\). Figure~\ref{fig:stdp_weight}b shows that the weights are indeed more staggered, but the classification accuracy decreases as \(\beta_+\) and \(\beta_-\) get larger (see Table~\ref{tab:snn_beta}). So, the fact that STDP leads to saturated weights is not the reason for the performance gap with AEs.

\begin{table}[ht]
\centering
\begin{tabular}{|c|c|c|c|c|}
\hline
Dataset & $\beta=1$ & $\beta=2$ & $\beta=3$ & $\beta=4$\\
\hline
CIFAR-10 & 48.27\(\pm\)0.47 & 46.56\(\pm\)0.68 & 43.18\(\pm\)1.60 & 41.03\(\pm\)0.21 \\
\hline
CIFAR-10-bw & 45.37\(\pm\)0.13 & 44.55\(\pm\)0.57 & 41.74\(\pm\)1.50 & 38.90\(\pm\)1.57 \\
\hline
\end{tabular}
\caption{SNN recognition rate according to STDP $\beta$ parameter (\(n_f=64\)).}
\label{tab:snn_beta}
\end{table}

Finally, the filters shown in Figure~\ref{figure:experiments:SNN_filters} also show a good property of SNNs: they do not raise any dead units, i.e. features that get stuck in a state with average weights that do not correspond to any significant pattern. By contrast, AEs tend to learn a fair amount of such features, especially when the number of features increases (see Figure~\ref{figure:experiments:AE_filters}). This behavior of SNNs can be due to two factors: lateral inhibition, which prevents neurons from learning similar patterns (here, becoming dead units), and the saturated regime of STDP.

\section{Result Analysis and Properties of the Networks}
\label{section:discussion}

\subsection{On-center/off-center coding}
In this section, we investigate the impact of on-center/off-center coding on classification accuracy. As mentioned in Section~\ref{section:experiments:snn_vs_ae}, this image coding is responsible for the type of visual features learned by STDP, but does it impact the final accuracy of the system? We compared the accuracy of two systems, each with and without preprocessing images: an AE, under the same protocol as before, and an SVM performing classification directly from image pixels. The AE parameters for the on-center/off-center coding runs are: \(\rho=0.005\), \(\gamma=1.0\), and \(\lambda=10^{-4}\). Results on CIFAR-10 and CIFAR-10-bw are reported in Table~\ref{tab:svm_raw}. Using on-center/off-center coding decreases the accuracy of the classification in both configurations, which confirms that this coding is one of the causes of the limited performance of SNNs in image classification. This is due to the fact that extracting edges with DoG has the effect of selecting only a subrange of frequencies. In addition, the accuracies obtained on filtered color images are only on par with (in the case of AEs) or worse than (with SVM) the results obtained using grayscale images; it highlights the fact that on-center/off-center coding cannot handle color effectively. One reason is that edge information is effectively represented by grayscale pixels, and the additional information brought by color is essentially located in uniform image regions. Interestingly, the unsupervised SNN models of the literature that are competitive with traditional approaches are only evaluated on the MNIST dataset~\cite{tavanaei18a}, which does not require on-center/off--center coding as the images are only made of edges (white handwritten digits on black backgrounds).

Therefore, to be effective, SNNs require the design of a suited image coding that preserves as much  visual information as possible. Using alternative methods to extract edges (such as the image gradient or the image Laplacian) could capture slightly different types of edge information, which could be processed within a single SNN for improved performance, in a feature fusion approach. However, this would only process edge information, which is insufficient to reach optimal classification performances. Ideally, SNNs should be able to handle raw RGB pixels; however, this is not straightforward, as we showed in Section~\ref{section:experiments:color}.

\begin{table}[ht]
\centering
\begin{tabular}{|l|c|c|}
\hline
Dataset & Raw pixels & AE features \(n_f=64\)\\
\hline
CIFAR-10 & 37.79 & 57.56\(\pm\)0.08\\
\hline
CIFAR-10-dog & 21.07 & 52.65\(\pm\)0.30\\
\hline
CIFAR-10-bw & 28.38 & 53.69\(\pm\)0.34\\
\hline
CIFAR-10-dog-bw & 25.29 & 52.76\(\pm\)0.08 \\
\hline
\end{tabular}
\caption{Classification accuracy (\%) obtained with raw pixels and AE features w.r.t. pre-processing methods. Only one run is performed on raw pixels as SVM training is deterministic.}
\label{tab:svm_raw}
\end{table}

\subsection{Sparsity}
We investigate here the sparsity properties of SNNs and AEs. To do so, we use the following sparseness measure~\cite{hoyer04a}:
\begin{equation}
\mathrm{sp}(f) = \frac{\sqrt{n_f} - \frac{\sum_i^{n_f} |f_i|}{\sqrt{\sum_i^{n_f} f_i^2}}}{\sqrt{n_f}-1}
\label{equation:sparseness}
\end{equation}
where \(f\) is the vector of activations of hidden units (i.e. the visual feature vector) and \(n_f\) is the number of hidden units. \(\mathrm{sp}(f) \in [0,1]\); larger values indicate sparser activations.

Table~\ref{table:discussion:sparseness} shows the mean sparseness of features computed on the test set of CIFAR-10. The sparseness is much higher in SNNs than in AEs. Indeed, the specialization of features in SNNs relies mostly on lateral inhibition, which prevents units from integrating spikes, leading to very sparse activations of the features. Sparsity is often cited as a necessary condition for good representations~\cite{bengio12a}, and has been shown to be correlated to classification accuracy on image datasets~\cite{jiang15a}. However, some results in~\cite{jiang15a} show that maximizing sparsity does not always lead to improvements in classification accuracy in AEs. Similarly, we observed experimentally that enforcing too much sparsity on the AEs (e.g., by lowering \(\rho\)) is detrimental to the classification accuracy. To push it further, we performed five runs on CIFAR-10 with \(n_f=64\) and different values for parameters \(\lambda\), \(\gamma\), and \(\rho\). The AE parameters were set so that the sparseness would be close to the sparseness that we measured in SNNs (i.e., in the range [0.8;0.9]). In these runs, the classification accuracy varies from 35.53\% to 41.03\%, much lower than our 57.56\% baseline. To check whether high levels of sparseness are an issue for SNNs too, we ran an experiment where lateral inhibition is deactivated during the feature extraction phase\footnote{Inhibition is still maintained during feature training, because SNNs cannot converge if there is no competition between neurons, as explained in Section~\ref{ssec:ihnibition}.}. 
As expected, deactivating inhibition decreased the sparseness of the model (from 0.869 to 0.638 on CIFAR-10). However, the classification rate decreased too (from  48.27\% to 47.35\%). It shows that, although sparsity is a desirable feature for good representations, an excessive level of sparseness can be detrimental, and that the right amount of sparsity should be enforced during training. This calls for the use of other, less restrictive, inhibition strategies than WTA.
\begin{table}[ht]
\centering
\begin{tabular}{|c|c|c|}
\hline
Model & \(n_f=64\) & \(n_f=1024\)\\
\hline
SNN & 0.869\(\pm 1.96 \mathrm{e}{-5}\) & 0.967\(\pm 3.04\mathrm{e}{-5}\)\\
\hline
AE & 0.352\(\pm\)0.116 & 0.112\(\pm\)0.077\\
\hline
\end{tabular}
\caption{Mean and standard deviation of feature sparseness (test set of CIFAR-10).}
\label{table:discussion:sparseness}
\end{table}

\subsection{Coherence}
One measure of the quality of the learned feature is their incoherence, i.e. the fact that one feature cannot be obtained by a sparse linear combination of other features in the vocabulary. If the incoherence is low, features are redundant, which is harmful for classification as redundant features will overweight other features. Inspired by the measure introduced in~\cite{makhzani14a}, we measure the coherence \(\mu_{ij}\) of two features \(f_i\) and \(f_j\) as their cosine similarity:
\begin{equation}
\mu_{ij} = \frac{|<f_i, f_j>|}{||f_i||_2.||f_j||_2}
\label{equation:coherence}
\end{equation}
where \(f_i\) is the \(i\)-th feature, \(<.,.>\) is the dot product operator, and \(||.||_2\) is the L2 vector norm. \(\mu \in [0,1]\); 0 corresponds to orthogonal (incoherent) features and 1 to similar (coherent) features. The weights span different ranges of values depending on the feature extractor considered; feature normalization makes coherence measures comparable from one extractor to another.

Table~\ref{table:discussion:coherence_average} displays the mean and the standard deviation of coherence measure \(\mu\) under all experimental settings. Overall, STDP-based SNNs produce more coherent features, which is one of the factors that can explain their lower performance. Moreover, the maximum pairwise coherence between two SNN feature is higher (\(\max(\mu_{ij}) = 0.999\) in most experimental configurations) than the maximum coherence between AE-produced features (\(\max(\mu_{ij}) \in [0.898, 0.998]\)), i.e. SNNs can learn almost identical features; in AEs, such features mostly correspond to dead units, whereas in SNNs they are significant features that are repeated. This result shows the limits of WTA inhibition, which should prevent features from reacting to the same patterns but fails to do so in practice. This calls for more work on understanding inhibition mechanisms and designing inhibition models that better prevent the co-adaptation of features.

\begin{table}[ht]
\centering
\begin{tabular}{|c|c|c|c|c|}
\hline
\multirow{2}{*}{Dataset} & \multicolumn{2}{c|}{SNN} & \multicolumn{2}{c|}{Autoencoder}\\
\cline{2-5}
& \(n_f=64\) & \(n_f=1024\) & \(n_f=64\) & \(n_f=1024\)\\
\hline
CIFAR-10 & 0.252\(\pm\)0.252 & 0.285\(\pm\)0.249 & 0.154\(\pm\)0.144 &  0.145\(\pm\)0.109 \\
\hline
CIFAR-10-bw & 0.313\(\pm\)0.271 & 0.340\(\pm\)0.234 & 0.119\(\pm\)0.138 & 0.225\(\pm\)0.161\\
\hline
CIFAR-100 & 0.256\(\pm\)0.230  & 0.289\(\pm\)0.238 & 0.154\(\pm\)0.149  &  0.143\(\pm\)0.199\\
\hline
CIFAR-100-bw & 0.320\(\pm\)0.238 & 0.343\(\pm\)0.223 & 0.121\(\pm\)0.137& 0.234\(\pm\)0.166\\
\hline
STL-10 & 0.263\(\pm\)0.293 & 0.293\(\pm\)0.246 & 0.177\(\pm\)0.164 & 0.151\(\pm\)0.114\\
\hline
STL-10-bw & 0.263\(\pm\)0.293 & 0.293\(\pm\)0.246 & 0.119\(\pm\)0.132 & 0.236\(\pm\)0.169\\
\hline
\end{tabular}
\caption{Mean and standard deviation of feature coherence \(\mu\) under all experimental settings.}
\label{table:discussion:coherence_average}
\end{table}

\subsection{Objective Function}
\begin{figure}[ht]
	\begin{center}
	\subfloat[]{
    	\begin{tabular}{cc}
    		\resizebox{0.15\textwidth}{!}{\includegraphics{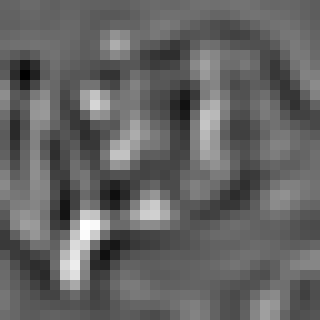}} \\
    		\resizebox{0.15\textwidth}{!}{\includegraphics{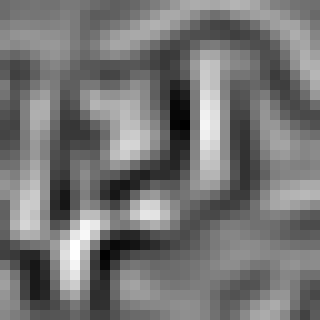}} \\
    	\end{tabular}
    }
	\subfloat[]{
    	\begin{tabular}{cc}
    		\resizebox{0.15\textwidth}{!}{\includegraphics{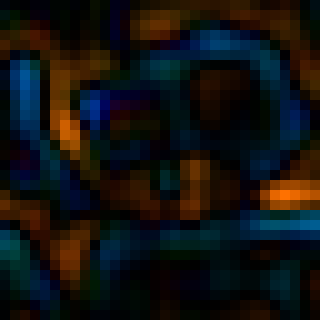}} \\
    		\resizebox{0.15\textwidth}{!}{\includegraphics{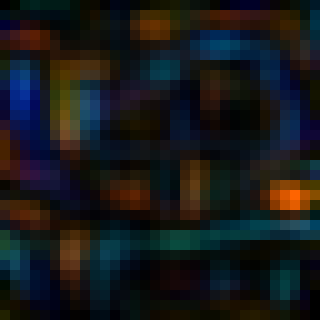}} \\
    	\end{tabular}
    }
    	\subfloat[]{
    	\begin{tabular}{cc}
    		\resizebox{0.15\textwidth}{!}{\includegraphics{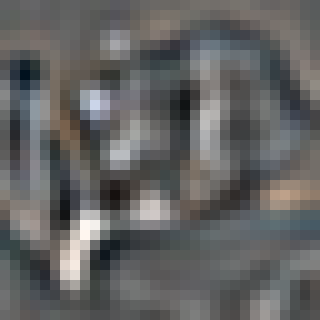}} \\
    		\resizebox{0.15\textwidth}{!}{\includegraphics{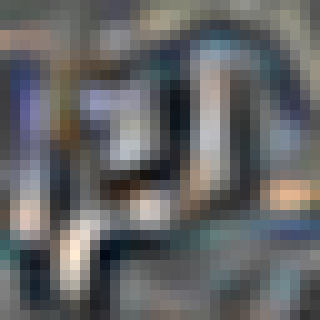}} \\
    	\end{tabular}
    }
	\subfloat[]{
    	\begin{tabular}{cc}
    		\resizebox{0.15\textwidth}{!}{\includegraphics{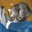}} \\
    		\resizebox{0.15\textwidth}{!}{\includegraphics{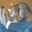}} \\
    	\end{tabular}
    }
	\subfloat[]{
    	\begin{tabular}{cc}
    		\resizebox{0.15\textwidth}{!}{\includegraphics{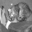}} \\
    		\resizebox{0.15\textwidth}{!}{\includegraphics{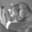}} \\
    	\end{tabular}
    }	\end{center}
    \caption{Image reconstruction samples from the test sets of CIFAR-10 and CIFAR-10-bw (top: pre-processed input images, bottom: reconstructed images). (a) SNN features, DoG-filtered grayscale image (b) SNN features, DoG-filtered color image (c) SNN features, grayscale and color DoG-filtered image (d) AE filters, color image (e) AE filters, grayscale image.}
	\label{figure:discussion:reconstruction_samples}
\end{figure}

\begin{figure}[ht]
	\begin{center}
	\subfloat[]{
    	\begin{tabular}{cc}
    		\resizebox{0.15\textwidth}{!}{\includegraphics{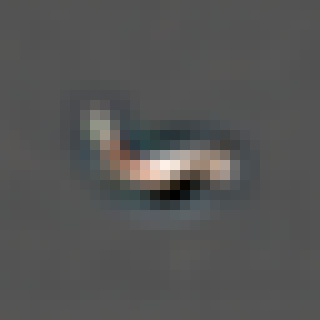}} &
    		\resizebox{0.15\textwidth}{!}{\includegraphics{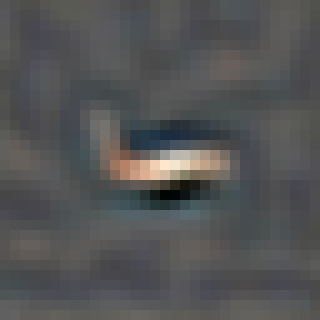}} \\
    	\end{tabular}
    }
    \hspace*{1cm}
	\subfloat[]{
    	\begin{tabular}{cc}
    		\resizebox{0.15\textwidth}{!}{\includegraphics{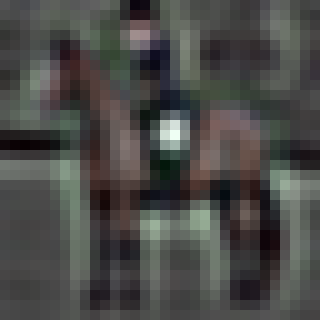}} &
    		\resizebox{0.15\textwidth}{!}{\includegraphics{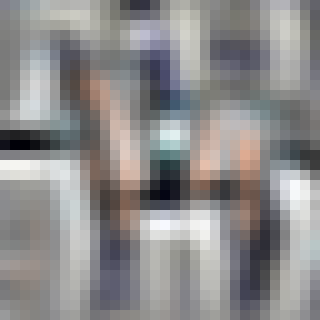}} \\
    	\end{tabular}
    }
	\end{center}
    \caption{(a) Best (error: 1.60) and (b) worst (error: 15.38) reconstructions from SNN features from the test set of CIFAR-10 (left: input images, right: reconstructions).}
	\label{figure:discussion:best_and_worst_reconstruction}
\end{figure}

One issue with STDP learning is that the objective function optimized by the system is not explicitly expressed, unlike AEs, which minimize reconstruction error. Identifying the criteria that are optimized by STDP rules would help better understand the related learning process and design learning rules for specific tasks. In this section, we check whether STDP rules embed reconstruction as a training criteria, by checking whether features learned through STDP are suited for image reconstruction, as those learned by AEs do.
To do so, we reconstructed the test images from the visual features. First, we reconstruct individual patches: in AEs, the reconstructed patches are directly provided by the decoder; in SNNs, patches are reconstructed as a linear combination of the filters weighted by their activations for the current sample, like in an AE with tied weights. Images are reconstructed from patches by averaging the values of overlapping patches at each location.
Table~\ref{table:discussion:reconstruction_error} shows the reconstruction error of each feature extractor on the test set of CIFAR-10, computed as the sum of squared errors between input images and reconstructed images, averaged over the samples. The reconstruction error is much higher for SNNs than AEs, which suggests that STDP does not learn features that allow reconstruction. However, qualitatively, the results look different (see samples in Figure~\ref{figure:discussion:reconstruction_samples}): the edges of the objects are reconstructed, although with less details than in the original images, but the global illumination is degraded. The degradation of pixel intensities explains for a large part the increased reconstruction error. This is best illustrated by the best and worst reconstructions (in the sense of the MSE) that we obtained using SNNs (see Figure~\ref{figure:discussion:best_and_worst_reconstruction}): edges are reconstructed correctly in both, but not pixel intensities. The reason for this is that SNNs process DoG-filtered images, in which color intensities are discarded and only edge information is retained. One could expect the reconstruction error of SNNs to be much lower if they were able to process raw images directly. Also, the lack of details around the edges could be blamed on the learned features being too elementary and sparse, which prevents the reconstruction of complex patterns.

These results show that, although this is not explicit in the learning rules, STDP learns to reconstruct images, among other potential criteria. However, it is known that minimizing reconstruction error is not sufficient to provide meaningful representations~\cite{vincent10a}. This is why recent AE models include additional criteria such has sparsity penalties~\cite{jiang15a} or Jacobian regularization~\cite{rifai11a}. How such criteria could be implemented within STDP rules, as well as which other criteria are already embedded in the STDP rules, are still open questions.

\begin{table}[ht]
\centering
\begin{tabular}{|c|c|c|c|c|}
\hline
 & \multicolumn{2}{|c}{SNN} & \multicolumn{2}{|c|}{AE}\\
\hline
 & \(n_f=64\) & \(n_f=1024\) & \(n_f=64\) & \(n_f=1024\)\\
\hline
CIFAR-10 & 4.9429 & 4.4179 & 0.0802 & 0.0742\\
\hline
CIFAR-10-bw & 4.9797 & 4.4628 & 0.00407 & 0.00472\\
\hline
\end{tabular}
\caption{Average reconstruction errors on the test set of CIFAR-10.}
\label{table:discussion:reconstruction_error}
\end{table}

\section{Conclusion}
\label{section:conclusion}
In this paper, we compared spiking neural networks (SNNs) equipped with STDP to auto-encoders (AEs) for unsupervised visual feature learning. Experiments on three image classification datasets showed that STDP cannot currently compete with classical neural networks trained with gradient descent, but also highlighted a number of properties of SNNs and provided specific directions towards effective feature learning with SNNs. Specifically, we showed that:
\begin{itemize}
	\item STDP-based SNNs are unable to deal naturally with RGB images;
	\item the common on-center/off-center image coding used in SNNs results in an information loss, thus harming the classification accuracy; this information loss is even more pronounced on color images;
    \item winner-take-all inhibition results in overly sparse features and does not prevent the co-adaptation of features in practice;
    \item STDP-based learning rules produce features that enable to reconstruct images from the learned features, as AEs do, even though the features are not explicitly optimized for this task.
\end{itemize}

These conclusions suggest two research directions to bridge the gap with standard neural networks. The first is to address the information loss caused by on-center/off-center coding and the conversion to grayscale images. One straightforward solution is to use standard pre-processing methods that have shown to be effective in computer vision, such as applying DoG filters in a scale-space fashion~\cite{lowe04a}, or replacing them with whitening~\cite{coates11a}, which highlights edges while retaining some color information. However, the resulting filters would still be limited to edge regions; dealing with non-edge information would require the design of alternatives to DoG filtering or new neural coding schemes that are not based on contrast only. The second direction is to ensure that the features have adequate amounts of sparsity and redundancy. To do so, one could design more suited inhibition mechanisms. Such mechanisms should be "soft", i.e. allow several neurons to fire at the same time, as demonstrated in~\cite{makhzani14a} in the case of AEs. The methods in~\cite{makhzani14a} could be good candidates but should be adapted to preserve the locality of computations. Another option is to design adaptive inhibition rules that maintain the homeostasis of the system w.r.t to some sparsity and co-adaptation objectives. Instead of inhibition rules, or in addition to them, learning rules with homeostatic plasticity~\cite{watt10a} can allow to reach the targeted sparsity and co-adaptation levels by acting on the synapses rather than the states of the neurons.

\section*{Acknowledgments}
This work was supported in part by IRCICA (Univ. Lille, CNRS,
USR 3380 -- IRCICA, F-59000 Lille, France) under the Bioinspired Project.

\section*{References}

\bibliographystyle{elsarticle-num-names}
\bibliography{main}

\end{document}

%% file: tikz/spiking_neuron.tikz
\begin{tikzpicture}[
	node distance=1mm,
	layer/.style={black, draw=black, dashed},
	neuron/.style={black, draw=black, circle, minimum size=0.75cm},
	neuron2/.style={black, draw=black, circle, minimum size=0.75cm,inner sep=0.5pt},
	synapse/.style={-{Latex}, shorten >=1pt, shorten <=1pt, >=stealth},
	spike/.pic={
    \draw () -- ++(#1,0) -- ++(0,0.2mm);
},
	spike2/.pic={
     \draw[black, thick] (0,0) -- ++(0,2mm);
}
]

	\node (layer_input_ref) {};

	\node (neuron_1_3)  at (layer_input_ref) {$I_3$};
	\node (neuron_1_2) [above = of neuron_1_3] {$I_2$};
	\node (neuron_1_1) [ above = of neuron_1_2] {$I_1$};
	\node (neuron_1_4) [ below = of neuron_1_3] {$I_4$};
	\node (neuron_1_5) [ below = of neuron_1_4] {$I_5$};
	\node (layer_1) [fit={(neuron_1_1)  (neuron_1_2) (neuron_1_3) (neuron_1_4) (neuron_1_5)}] {};

	\node (neuron) [neuron2, right = 40mm of layer_1]  {$N$};

	\node (output_layer)  [right = 40mm of neuron] {};

	\node (neuron_2_2)  at (output_layer) {$O_2$};
	\node (neuron_2_1) [above = of neuron_2_2] {$O_2$};
	\node (neuron_2_3) [ below = of neuron_2_2] {$O_1$};

	\coordinate (input_end) at ($(neuron)+(-10mm,0)$);

	\draw [synapse] (neuron_1_1)  -- (neuron_1_1 -| input_end) -- (neuron);
	\draw [synapse] (neuron_1_2)  -- (neuron_1_2 -| input_end) -- (neuron);
	\draw [synapse] (neuron_1_3)  -- (neuron_1_3 -| input_end) -- (neuron);
	\draw [synapse] (neuron_1_4)  -- (neuron_1_4 -| input_end) -- (neuron);
	\draw [synapse] (neuron_1_5)  -- (neuron_1_5 -| input_end) -- (neuron);

	\coordinate (output_end) at ($(output_layer)+(-10mm,0)$);
	
	\draw(neuron) --  (output_end);
	\draw[synapse] (output_end) --  (neuron_2_1);
	\draw[synapse] (output_end) --  (neuron_2_2);
	\draw[synapse] (output_end) --  (neuron_2_3);

	\pic at  ($(neuron_1_1)+(10mm,0)$) {spike2};
	\pic at  ($(neuron_1_1)+(13mm,0)$) {spike2};
	\pic at  ($(neuron_1_1)+(26mm,0)$) {spike2};

	\pic at   ($(neuron_1_2)+(20mm,0)$) {spike2};

	\pic at ($(neuron_1_4)+(5mm,0)$){spike2};
	\pic at ($(neuron_1_4)+(12mm,0)$) {spike2};
	\pic at ($(neuron_1_4)+(28mm,0)$) {spike2};

	\pic at ($(neuron_1_5)+(20mm,0)$) {spike2};
	\pic at ($(neuron_1_5)+(23mm,0)$){spike2};

	\pic at  ($(neuron)+(12mm,0)$) {spike2};
	\pic at   ($(neuron)+(22mm,0)$) {spike2};

	\coordinate(axis) at(0,-20mm);
	\draw[-{Latex}] (layer_1.east |- axis) -- node[below]{t} (input_end|- axis);
	\draw[-{Latex}] (neuron.east |- axis) -- node[below]{t} (output_end|- axis);

\end{tikzpicture}

%% file: tikz/if.tikz
\begin{tikzpicture}

\begin{axis}[
	use fpu=false,
	name=main,
	height=8cm, width=10cm, xmin=0, xmax=100, ymin=-1, ymax=3, 
	xlabel=$t$ (ms),
   	ylabel=$v$ (mv),
	samples=50,
    legend pos = outer north east
]

  	\addplot[domain=0:100,black, dashed, forget plot] (x, 2);
	\addplot[domain=0:100,black, forget plot] (x, 0);

	\addplot[domain=0:10,blue, ultra thick, forget plot] (x, 0);
 	\addplot[mark=none,blue, ultra thick, forget plot] coordinates {(10,0) (10,0.8)};
	\addplot[domain=10:30,blue, ultra thick, forget plot] (x, 0.8);
	\addplot[mark=none, blue, ultra thick, forget plot] coordinates {(30,0.8) (30,1.2)};
	\addplot[domain=30:35,blue, ultra thick, forget plot] (x, 1.2);
	\addplot[mark=none,blue, ultra thick, forget plot] coordinates {(35,1.2) (35,1.9)};
	\addplot[domain=35:37,blue, ultra thick, forget plot] (x, 1.9);
	\addplot[mark=none,blue, ultra thick, forget plot] coordinates {(37,0) (37,2.2)};
	\addplot[domain=37:47,red, ultra thick, forget plot] (x, 0);
	\addplot[domain=47:65, blue, ultra thick, forget plot] (x, 0);
	\addplot[mark=none,blue, ultra thick, forget plot] coordinates {(65,0) (65,0.5)};
	\addplot[domain=65:100, blue, ultra thick, forget plot] (x, 0.5);

    \addlegendimage{line legend,blue,ultra thick}
    \addlegendentry{Membrane potential}
    \addlegendimage{line legend,red,ultra thick}
    \addlegendentry{Refractory period}
    \addlegendimage{line legend,black,dashed}
    \addlegendentry{Membrane potential threshold}
    \addlegendimage{only marks, mark=*, blue}
    \addlegendentry{Incoming spikes}
    \addlegendimage{only marks, mark=*, red}
    \addlegendentry{Outgoing spikes}
\end{axis}

\begin{axis}[
	height=2cm, width=10cm, xmin=0, xmax=100, ymin=-1, ymax=1, 
	at=(main.below south), anchor=above north,
	yticklabels=\empty,
	major tick length=0pt,
    minor tick length=0pt,
	samples=50,
]
	\addplot[mark=*, blue, forget plot] coordinates {(10,-0.5)};
	\addplot[mark=*, blue, forget plot] coordinates {(30,-0.5)};
	\addplot[mark=*, blue, forget plot] coordinates {(35,-0.5)};
	\addplot[mark=*, blue, forget plot] coordinates {(37,-0.5)};
	\addplot[mark=*, red, forget plot] coordinates {(37,0.5)};
	\addplot[mark=*, blue, forget plot] coordinates {(42,-0.5)};

	\addplot[mark=*, blue, forget plot] coordinates {(65,-0.5)};

\end{axis}
\end{tikzpicture}

%% file: tikz/bio_stdp.tikz
\begin{tikzpicture}
\begin{axis}[xmin=-50, xmax=100,ymin=-0.3,ymax=0.3, 
	xlabel= $\Delta t$ (ms),
   	ylabel=$\Delta w$,
	samples=50]
  \addplot[domain=-50:-0.1, blue, ultra thick] {0.3125*e^(x/(16.8)};
  \addplot[domain=0.1:100,red,  ultra thick] {-0.85*0.3125*e^(-x/33.7)};
  \addplot[domain=-50:100,black, dashed] (x, 0);
  \addplot[black, dashed] (0, x);
  \addlegendentry{LTP}
  \addlegendentry{LTD}
\end{axis}
\end{tikzpicture}

%% file: tikz/simple_stdp.tikz
\begin{tikzpicture}
\begin{axis}[xmin=-100, xmax=50,ymin=-0.2,ymax=0.2, 
	xlabel= $\Delta t$ (ms),
   	ylabel=$\Delta w$,
	samples=50]
  \addplot[domain=-50:0, blue, ultra thick] (x, 0.1);
  \addplot[domain=-100:-50,red,  ultra thick] (x, -0.05);
  \addplot[domain=0:50,red,  ultra thick] (x, -0.05);
  \addplot[domain=-100:50,black, dashed] (x, 0);
  \addplot[black, dashed] (0, x);
  \addlegendentry{LTP}
  \addlegendentry{LTD}
\end{axis}
\end{tikzpicture}

%% file: tikz/coding.tikz
\begin{tikzpicture}

\draw[->] (0,0) -- (8,0) node[anchor=north] {\Large t};

\draw[dashed] (1, 0) -- (1, 6);
\draw[dashed] (4, 0) -- (4, 6);
\draw[dashed] (7, 0) -- (7, 6);

\node at (2.5, -0.5) {\Large Sample 1};
\node at (5.5, -0.5) {\Large Sample 2};

\node at (0, 1) {\Large $I_1$};
\node at (0, 3) {\Large $I_2$};
\node at (0, 5) {\Large $I_3$};

\node[fill, circle,minimum size=2mm, inner sep=0pt] at (1.2, 5) {};
\node at (2.5, 4.5) {\large $x_3 = 0.9$};

\node[fill, circle,minimum size=2mm, inner sep=0pt] at (5.6, 5) {};
\node at (5.5, 4.5) {\large $x_3 = 0.48$};

\node at (2.5, 2.5) {\large $x_2 = 0.0$};

\node[fill, circle,minimum size=2mm, inner sep=0pt] at (6.8, 3) {};
\node at (5.5, 2.5) {\large $x_2 = 0.1$};

\node[fill, circle,minimum size=2mm, inner sep=0pt] at (2, 1) {};
\node at (2.5, 0.5) {\large $x_1 = 0.7$};

\node[fill, circle,minimum size=2mm, inner sep=0pt] at (6, 1) {};
\node at (5.5, 0.5) {\large $x_1 = 0.3$};
\end{tikzpicture}

%% file: tikz/topology.tikz
\begin{tikzpicture}[scale=.9,every node/.style={minimum size=1cm},on grid]
		
    \begin{scope}[
            ,every node/.append style={
            yslant=0,xslant=0},yslant=0.5,xslant=0.0,xscale=0.8
            ]
        \fill[white,fill opacity=0.9] (0,0) rectangle (5,5);
        \draw[step=10mm, black] (0,0) grid (5,5); 

        \draw[black,very thick] (0,0) rectangle (5,5);

	\node at(-0.5, 2.5) {\Large $w_p$};
	\node at(2.5, 5.5) {\Large $w_p$};

	\coordinate (o1) at (4.7,-1.1);
	\coordinate (o2) at (4.7,-0);
	\coordinate (o3) at (10.3,-1.1);
	\coordinate (o4) at (10.3,-0);

	\draw[dashed] (0,0) -- (o1) ;
	\draw[dashed] (0,5) -- (o2) ;
	\draw[dashed] (5,0) -- (o3) ;
	\draw[dashed] (5,5) -- (o4) ;

	\draw[dashed] (o1) -- (o3) ;
	\draw[dashed] (o2) -- (o4) ;
	\draw[dashed] (o3) -- (o4) ;
	\draw[dashed] (o1) -- (o2) ;

    \end{scope}

    \begin{scope}[
           xshift=120,yshift=-30,every node/.append style={
            yslant=0,xslant=0},yslant=0.5,xslant=0.0,xscale=1
            ]

	\node[fill= white, draw=black, circle, minimum size=0.8cm] at(0, 2.5) {};
	\node[fill= white, draw=black, circle, minimum size=0.8cm] at(1, 2.5) {};
	\node[fill= white, draw=black, circle, minimum size=0.8cm] at(2, 2.5) {};
	\node[fill= white, draw=black, circle, minimum size=0.8cm] at(3, 2.5) {};
	\node[fill= white, draw=black, circle, minimum size=0.8cm] at(4, 2.5) {};
	
	\node at(1.8, 3.5) {\Large $n_f$};

	\draw[<->,xshift=7,yshift=-15] (0, 2.5) edge[bend right = 50] (1, 2.5) ;
	\draw[<->,xshift=7,yshift=-15] (0, 2.5) edge[bend right = 50] (2, 2.5) ;
	\draw[<->,xshift=7,yshift=-15] (0, 2.5) edge[bend right = 50] (3, 2.5) ;
	\draw[<->,xshift=7,yshift=-15] (0, 2.5) edge[bend right = 50] (4, 2.5) ;

	\draw[<->,xshift=7,yshift=-15] (1, 2.5) edge[bend right = 50] (2, 2.5) ;
	\draw[<->,xshift=7,yshift=-15] (1, 2.5) edge[bend right = 50] (3, 2.5) ;
	\draw[<->,xshift=7,yshift=-15] (1, 2.5) edge[bend right = 50] (4, 2.5) ;

	\draw[<->,xshift=7,yshift=-15] (2, 2.5) edge[bend right = 50] (3, 2.5) ;
	\draw[<->,xshift=7,yshift=-15] (2, 2.5) edge[bend right = 50] (4, 2.5) ;
	
	\draw[<->,xshift=7,yshift=-15] (3, 2.5) edge[bend right = 50] (4, 2.5) ;
    \end{scope}

\end{tikzpicture}

%% file: tikz/ae.tikz
\begin{tikzpicture}[scale=.9,every node/.style={minimum size=1cm},on grid]
		
    \begin{scope}[
            ,every node/.append style={
            yslant=0,xslant=0},yslant=0.5,xslant=0.0,xscale=0.8
            ]
        \fill[white,fill opacity=0.9] (0,0) rectangle (5,5);
        \draw[step=10mm, black] (0,0) grid (5,5); 

	  \draw[step=10mm, black] (0,0) grid (5,5); 

        \draw[black,very thick] (0,0) rectangle (5,5);

	\node at(-0.5, 2.5) {\Large $w_p$};
	\node at(2.5, 5.5) {\Large $w_p$};

	\coordinate (o1) at (4.7,-1.1);
	\coordinate (o2) at (4.7,-0);
	\coordinate (o3) at (10.3,-1.1);
	\coordinate (o4) at (10.3,-0);

	\draw[dashed] (0,0) -- (o1) ;
	\draw[dashed] (0,5) -- (o2) ;
	\draw[dashed] (5,0) -- (o3) ;
	\draw[dashed] (5,5) -- (o4) ;

	\draw[dashed] (o1) -- (o3) ;
	\draw[dashed] (o2) -- (o4) ;
	\draw[dashed] (o3) -- (o4) ;
	\draw[dashed] (o1) -- (o2) ;

    \end{scope}

    \begin{scope}[
           xshift=260,yshift=-50,every node/.append style={
            yslant=0,xslant=0},yslant=0.5,xslant=0.0,xscale=0.8
            ]

	\coordinate (o4) at (-1.1,6.32);
	\coordinate (o3) at (-1,5.1);

	\draw[dashed] (5,0) -- (o3) ;
	\draw[dashed] (5,5) -- (o4) ;

	 \end{scope}

    \begin{scope}[
           xshift=120,yshift=-30,every node/.append style={
            yslant=0,xslant=0},yslant=0.5,xslant=0.0,xscale=1
            ]

	\node[fill= white, draw=black, circle, minimum size=0.8cm] at(0, 2.5) {};
	\node[fill= white, draw=black, circle, minimum size=0.8cm] at(1, 2.5) {};
	\node[fill= white, draw=black, circle, minimum size=0.8cm] at(2, 2.5) {};
	\node[fill= white, draw=black, circle, minimum size=0.8cm] at(3, 2.5) {};
	\node[fill= white, draw=black, circle, minimum size=0.8cm] at(4, 2.5) {};
	
	\node at(1.8, 3.5) {\Large $n_f$};

    \end{scope}

    \begin{scope}[
           xshift=260,yshift=-50,every node/.append style={
            yslant=0,xslant=0},yslant=0.5,xslant=0.0,xscale=0.8
            ]
	\coordinate (o2) at (-6.8,6.3);
	\coordinate (o1) at (-6.75,5.25);
	\draw[dashed] (0,0) -- (o1) ;
	\draw[dashed] (0,5) -- (o2) ;

        \fill[white,fill opacity=0.9] (0,0) rectangle (5,5);
        \draw[step=10mm, black] (0,0) grid (5,5); 
	 \draw[black,very thick] (0,0) rectangle (5,5);
	\node at(-0.5, 2.5) {\Large $w_p$};
	\node at(2.5, 5.5) {\Large $w_p$};

	 \end{scope}
\end{tikzpicture}

%% file: tikz/classification.tikz
\usetikzlibrary{calc}

\newcommand{\rectangle}[4]
{ 
	\draw[fill=white] (#1,#2) -- (#1+#3,#2) -- (#1+#3,#2+#4) -- (#1,#2+#4) --  (#1,#2) ;
}

\newcommand{\dashedrectangle}[5]
{ 
	\draw[fill=white, dashed,opacity=#5] (#1,#2) -- (#1+#3,#2) -- (#1+#3,#2+#4) -- (#1,#2+#4) --  (#1,#2) ;
}

\begin{tikzpicture}

	\def\imagex{0}
	\def\imagey{0}
	\def\imagew{4}
	\def\imageh{4}

	\def\patchi{1.5}
	\def\patchix{0.7}
	\def\patchiy{0.6}
	\def\patchis{0.4}

	\def\convx{6}
	\def\convy{0.3}
	\def\convw{3}
	\def\convh{3}

	\def\patchc{0.3}
	\def\patchcx{0.4}
	\def\patchcy{0.4}

   \def\patchd{1.5}
	\def\patchdx{1.4}
	\def\patchdy{1.4}

	\def\poolx{11}
	\def\pooly{0}
	\def\poolw{0.4}
	\def\poolh{4}

   \def\patchpw{0.3}
   \def\patchph{0.9}
	\def\patchpx{0.05}
	\def\patchpy{3.05}

	\def\depthf{0.1}

	\rectangle{\imagex+0*\depthf}{\imagey+0*\depthf}{\imagew}{\imageh}

	
	\dashedrectangle{\imagex+\patchix+1*\patchis}{\imagey+\imageh-\patchi-\patchiy-1*\patchis}{\patchi}{\patchi}{0.2}
	\dashedrectangle{\imagex+\patchix+0*\patchis}{\imagey+\imageh-\patchi-\patchiy-2*\patchis}{\patchi}{\patchi}{0.2}
	\dashedrectangle{\imagex+\patchix+2*\patchis}{\imagey+\imageh-\patchi-\patchiy-0*\patchis}{\patchi}{\patchi}{0.2}
	\dashedrectangle{\imagex+\patchix+0*\patchis}{\imagey+\imageh-\patchi-\patchiy-1*\patchis}{\patchi}{\patchi}{0.5}
	\dashedrectangle{\imagex+\patchix+1*\patchis}{\imagey+\imageh-\patchi-\patchiy-0*\patchis}{\patchi}{\patchi}{0.5}
	\dashedrectangle{\imagex+\patchix+0*\patchis}{\imagey+\imageh-\patchi-\patchiy-0*\patchis}{\patchi}{\patchi}{0.9}


	\node at (\imagex+\patchix-0.4, \imagey+\imageh-\patchiy-\patchi/2) {\large $w_p$};
	\node at (\imagex+\patchix+\patchi/2, \imagey+\imageh-\patchiy+0.3) {\large $w_p$};

	\node at (\imagex+\patchix-0.4, \imagey+\imageh-\patchiy-\patchi-\patchis/2) {\large $s$};

	\node at (5, 5) {\Large Feature extraction};

	\node at (\imagex+\imagew/2, -0.8) {\Large (a)};

	\rectangle{\convx+3*\depthf}{\convy+3*\depthf}{\convw}{\convh}
	\rectangle{\convx+2*\depthf}{\convy+2*\depthf}{\convw}{\convh}
	\rectangle{\convx+1*\depthf}{\convy+1*\depthf}{\convw}{\convh}
	\rectangle{\convx+0*\depthf}{\convy+0*\depthf}{\convw}{\convh}

	\dashedrectangle{\convx+\patchcx}{\convy+\convh-\patchc-\patchcy}{\patchc}{\patchc}{0.9}
	\draw[dotted] (\imagex+\patchi+\patchix, \imagey+\imageh-\patchiy) -- (\convx+\patchcx, \convy+\convh-\patchcy);
	\draw[dotted] (\imagex+\patchi+\patchix, \imagey+\imageh-\patchiy-\patchi) -- (\convx+\patchcx, \convy+\convh-\patchcy-\patchc);

	\node at (\convx-0.4, \convy+\convh/2) {\large $k$};
	\node at (\convx+\convw/2, \convy-0.4) {\large $k$};

	\node [rotate=50+90+180] at (\convx, \convy+\convh+0.2) {\huge $\left\{ \right.$};
	\node at (\convx-0.3, \convy+\convh+0.4) {\large $n_f$};

	\node at (9.8, 5) {\Large Sum pooling};
	\node at (\convx+\convw/2, -0.8) {\Large (b)};
	\rectangle{\poolx+0*\depthf}{\pooly+0*\depthf}{\poolw}{\poolh}

	\dashedrectangle{\convx+\patchdx}{\convy+\convh-\patchd-\patchdy}{\patchd}{\patchd}{0.9}
	\dashedrectangle{\poolx+\patchpx}{\pooly+\poolh-\patchph-\patchpy}{\patchpw}{\patchph}{0.9}
	\draw[dotted] (\convx+\patchdx+\patchd,\convy+\convh-\patchd-\patchdy) -- (\poolx+\patchpx, \pooly+\poolh-\patchph-\patchpy);
	\draw[dotted] (\convx+\patchdx+\patchd,\convy+\convh-\patchd-\patchdy+\patchd) -- (\poolx+\patchpx, \pooly+\poolh-\patchph-\patchpy+\patchph);

	\node[rotate=90] at (\poolx-0.3, \pooly+\poolh/2) {\large $r \times r \times n_f$};
    
  \node [rotate=180] at (\poolx+\poolw+0.3, \pooly+\poolh-\patchpy-\patchph/2) {\huge $\left\{ \right.$};
  
    \node at (\poolx+\poolw+0.9, \pooly+\poolh-\patchpy-\patchph/2-0.1) {\large $n_f$};

	\draw[->] (\poolx+\poolw+0.5,\pooly+\poolh/2) -- (\poolx+\poolw+1.2,\pooly+\poolh/2);
	\node at (\poolx+\poolw+2,\pooly+\poolh/2) {\Large SVM};
		\node at (\poolx+\poolw/2, -0.8) {\Large (c)};
	
\end{tikzpicture}

%% file: tikz/w_histo.tikz
\pgfplotstableread[col sep=space]{
0.025 2114
0.075 21
0.125 16
0.175 16
0.225 28
0.275 21
0.325 13
0.375 21
0.425 17
0.475 21
0.525 18
0.575 17
0.625 27
0.675 15
0.725 16
0.775 13
0.825 12
0.875 10
0.925 18
0.975 185
}\data

\begin{tikzpicture}
\begin{axis}[
	ybar,
    ymin=1,
    ymax=2500,
	ylabel=Population,
    xlabel=W,
	ymode=log,
    bar width=4pt,
    enlarge x limits={abs=5mm}
]
\addplot[fill, blue] table [x, y] {\data};
\end{axis}
\end{tikzpicture}

%% file: tikz/w_histo2.tikz
\pgfplotstableread[col sep=space]{
0.025 1107
0.075 235
0.125 188
0.175 178
0.225 144
0.275 94
0.325 80
0.375 80
0.425 62
0.475 53
0.525 40
0.575 52
0.625 67
0.675 102
0.725 106
0.775 121
0.825 140
0.875 121
0.925 91
0.975 70
}\data

\begin{tikzpicture}
\begin{axis}[
	ybar,
    ymin=1,
    ymax=2500,
	ylabel=Population,
    xlabel=W,
	ymode=log,
    bar width=4pt,
    enlarge x limits={abs=5mm}
]
\addplot[fill, blue] table [x, y] {\data};
\end{axis}
\end{tikzpicture}